\documentclass[sigconf,10pt]{acmart}

\AtBeginDocument{%
  \providecommand\BibTeX{{%
    Bib\TeX}}}

\makeatletter
\patchcmd{\maketitle}{\@copyrightspace}{}{}{}
\makeatother

\usepackage{amsmath,amsfonts}
\usepackage{url}
\usepackage{verbatim}
\usepackage{graphicx}
\usepackage{subfigure}
\usepackage{caption}

\usepackage{algorithm}
\usepackage{algorithmic}
\usepackage{mdwlist}
\usepackage{textcomp}
\usepackage{multirow}
\usepackage{url}
\usepackage{color}
\usepackage{pifont}
\usepackage{xcolor}
\usepackage{tabularx}
\usepackage{pifont}
\usepackage{booktabs}
\usepackage{makecell}
\usepackage{booktabs}
\usepackage{hyperref}
\usepackage[most]{tcolorbox}
\usepackage{subcaption}

\newcounter{glossy_enum}
\makecompactlist{list*}{list}

\newenvironment{glossy_itemize}
{\begin{list*}[{\labelitemi}{\topsep=0.2em \leftmargin=1.4em \itemindent=-0.2em}]}
{\end{list*}\vspace*{0.5em}}

\tcbset{
  myobservation/.style={
    enhanced,
    colback=gray!10!white,       % 浅灰色背景
    colframe=gray!50!black,      % 边框颜色
    arc=2mm,                     % 圆角半径
    boxsep=0pt,                  % 减少内容与边框的总体间距
    left=3mm,                    % 减少左边距
    right=3mm,                   % 减少右边距
    top=1.5mm,                   % 减少上边距
    bottom=1.5mm,                % 减少下边距
    before upper=\vspace{-1mm},  % 减少标题与内容的间距
    fonttitle=\normalfont\bfseries, % 标题使用正常字体但加粗
    coltitle=black,              % 标题颜色设为黑色（关键修改）
    boxrule=0.5pt,               % 边框宽度
    drop shadow,                 % 阴影效果
    fonttitle=\bfseries,         % 标题加粗
    title=Observation #1,        % 自动编号标题
    attach title to upper={\quad} % 标题与内容间距
  }
}

\newcommand{\fakepar}[1]{\vspace{.5mm}\noindent\textbf{#1.}}

\AtBeginDocument{%
  \providecommand\BibTeX{{%
    \normalfont B\kern-0.5em{\scshape i\kern-0.25em b}\kern-0.8em\TeX}}}

\setcopyright{rightsretained}

\begin{document}
% \begin{sloppypar}
\title{DynaFilter: Cloud-driven Dynamic Filtering for Satellite Edge Intelligence}

\author{Ziyang Zhang}
\affiliation{%
  \institution{Politecnico di Milano}
  \country{Milan, Italy}}
\email{ziyang.zhang@polimi.it}

\author{Jie Liu}
\affiliation{%
  \institution{Harbin Institute of Technology}
  \country{Shenzhen, China}}
\email{jieliu@hit.edu.cn}

\author{Luca Mottola}
\affiliation{%
  \institution{Politecnico di Milano}
  \country{Milan, Italy}}
\email{luca.mottola@polimi.it}

\renewcommand{\shortauthors}{Zhang et al.}

\begin{abstract}
Modern satellite edge systems, including those performing remote sensing tasks such object detection and tracking, are characterized by severely limited bandwidth and intermittent connections, making continuous data transmission to the cloud impractical. 
Existing edge-cloud systems, however, either require heavy pre-processing before analysis, for instance, full decompression of imagery data, or transmit all compressed data regardless of relevance. 
To address these challenges, we design DynaFilter, a dynamic filtering technique that enables satellite edge devices to perform selective region-of-interest (RoI) inference directly in the compressed-domain, without full decompression.
Our key insight is that low-level compression syntax, specifically DC coefficients/AC energy in JPEG images and motion vectors  in video streams, exhibits strong correlations with high-level semantic queries. 
By establishing a precise mapping between cloud query semantics and multimodal compressed-domain features, DynaFilter enables the edge to identify and transmit only relevant data associated to RoIs.
Extensive evaluations show that DynaFilter reduces the total volume of pixel data for decoding and subsequent inference by $1.6\times$$\sim$$7.1\times$ for images, and achieves 92.0\% bandwidth savings for video streams compared to state-of-the-art baselines. Furthermore, it decreases energy consumption by 43.1$\sim$88.6\% on target devices and achieves a $1.6\times$$\sim$$3.0\times$ speedup in inference latency.
\end{abstract}


%    <concept>
%        <concept_id>10003120.10003138</concept_id>
%        <concept_desc>Human-centered computing~Ubiquitous and mobile computing</concept_desc>
%        <concept_significance>300</concept_significance>
%        </concept>
%  </ccs2012>

% \ccsdesc[300]{Human-centered computing~Ubiquitous and mobile computing}




% \keywords{edge intelligence, cloud-driven, dynamic filtering, DNN inference}

%% These commands are for a PROCEEDINGS abstract or paper.

\copyrightyear{2026}
\acmYear{2026}
\setcopyright{acmlicensed}\acmConference[ACM MobiCom '26]{The 32nd Annual International Conference on Mobile Computing and Networking}{October 26--30,2026}{Austin, Texas, USA}
\acmBooktitle{The 32nd Annual International Conference on Mobile Computing and Networking (ACM MobiCom '26), October 26--30, 2026, Austin, Texas, USA}
\acmPrice{15.00}
\acmDOI{10.1145/3583120.3586953}
\acmISBN{979-8-4007-0118-4/23/05}

% \copyrightyear{2018}
% \acmYear{2018}
% \acmDOI{XXXXXXX.XXXXXXX}
% %% These commands are for a PROCEEDINGS abstract or paper.
% \acmConference[Conference acronym 'XX]{Make sure to enter the correct
%   conference title from your rights confirmation email}{June 03--05,
%   2018}{Woodstock, NY}
% %%
% %%  Uncomment \acmBooktitle if the title of the proceedings is different
% %%  from ``Proceedings of ...''!
% %%
% %%\acmBooktitle{Woodstock '18: ACM Symposium on Neural Gaze Detection,
% %%  June 03--05, 2018, Woodstock, NY}
% \acmISBN{978-1-4503-XXXX-X/2018/06}

\maketitle

\section{Introduction}\label{sec:introduction} 
Edge computing is making it into space systems, especially those built with off-the-shelf hardware and thus inherently resource constrained~\cite{orbitalEdge,denby2019orbital,berck2025launch,narayana2020hummingbird}. 
This evolution is particularly transformative for remote sensing tasks, such as environmental monitoring~\cite{shenoy2024cosmac} and aerial surveillance~\cite{tao2023transmitting}.

\begin{figure}[tb]
\vspace{-0.3cm}
\setlength{\abovecaptionskip}{0pt}
\setlength{\belowcaptionskip}{0pt}
\centering
\subfigure[DOTA-v1.0 Dataset]{
\begin{minipage}[b]{0.4875\linewidth}
\includegraphics[width=1\linewidth]{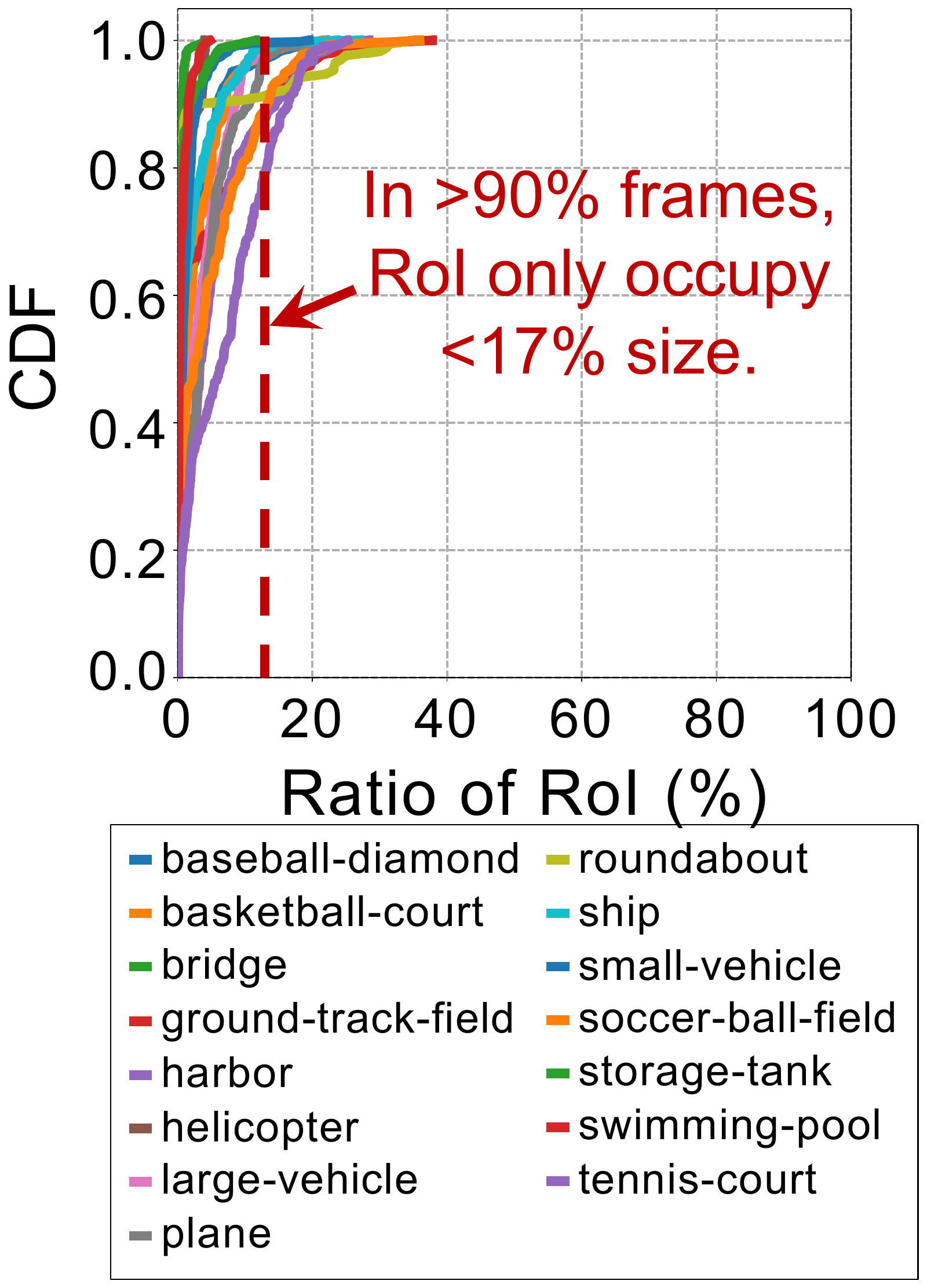}
\end{minipage}}
\subfigure[VisDrone 2019 Dataset]{
\begin{minipage}[b]{0.4825\linewidth}
\includegraphics[width=.99\linewidth]{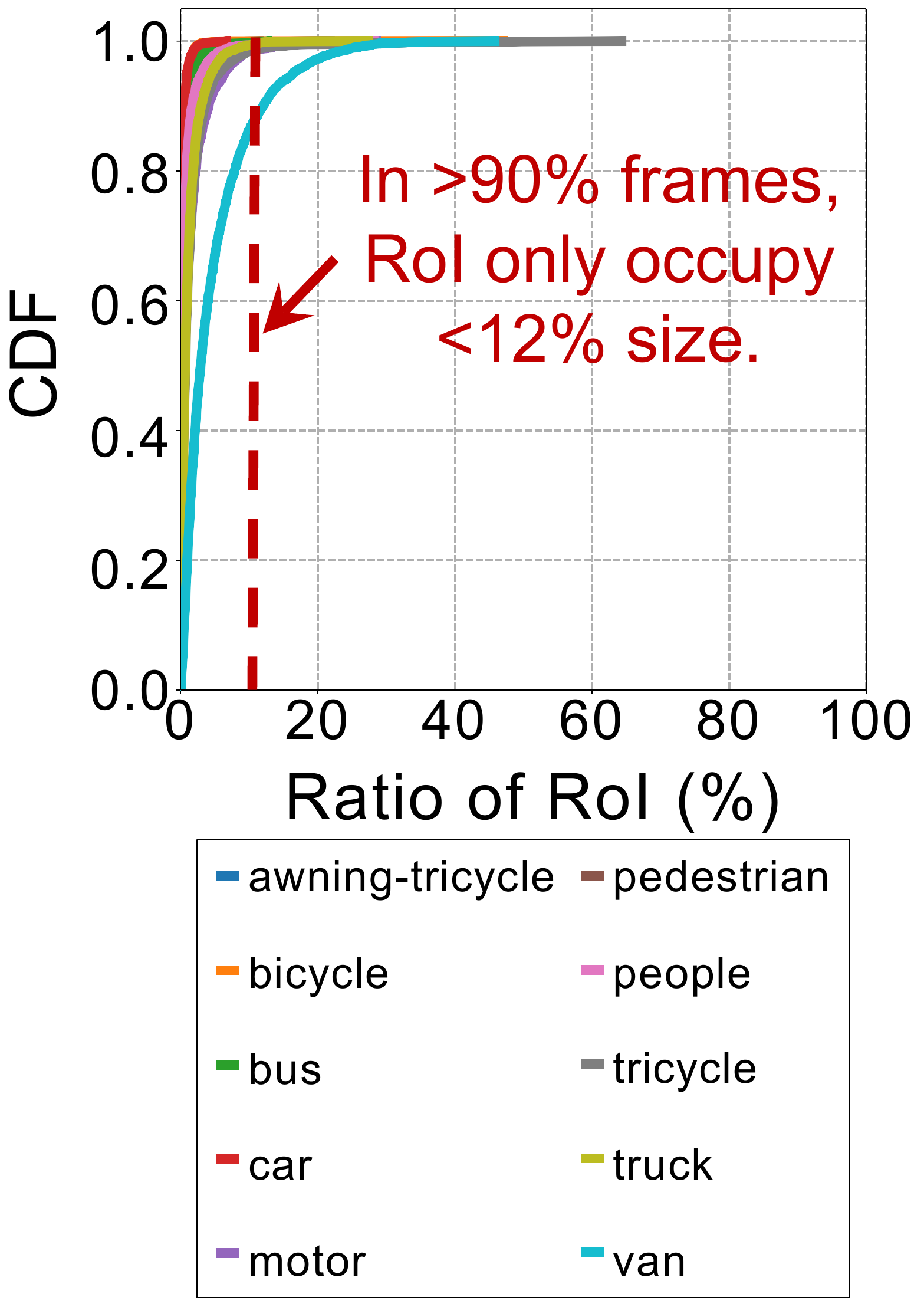}
\end{minipage}}
% \vspace{-0.5cm}
\caption{In over 90\% of frames over DOTA-V1.0 and VisDrone 2019 datasets, the region of interest (RoI) for object detection tasks only occupies 12$\sim$17\% of the spatial area of a frame.}
\Description{In over 90\% of frames over DOTA-V1.0 and VisDrone 2019 datasets, RoI (i.e., the ground reality bounding box) for object detection task only occupy 12$\sim$17\% of the spatial area of a frame.}
\vspace{-0.5cm}
\label{fig:observation}
\end{figure}

\fakepar{Satellite edge computing} In modern satellite edge computing architectures~\cite{tao2024known,denby2019orbital,orbitalEdge}, visual data captured by orbiting devices is typically compressed using standard codecs like JPEG or H.264 before transmission to cloud servers on the ground for analysis.
However, most existing systems transmit all compressed data regardless of relevance~\cite{li2021appealnet,banitalebi2021auto}.
This is fundamentally impractical in severely bandwidth-constrained environments, or in the presence of intermittent connections, as in most modern satellite systems, especially those built with off-the-shelf hardware or exploiting opportunistic connections to the ground~\cite{orbitalEdge,denby2019orbital,berck2025launch,yaacoub2025fault,satnogs}.

The fundamental challenge arises because edge devices must transmit compressed data through constrained channels, often providing kbps of bandwidth, if at all~\cite{lu2020edge,matsubara2022supervised,chen2022update}, while the cloud's analytical requirements, such as identifying specific object categories with minimum confidence thresholds, dynamically evolve based on mission objectives. Thus, edge devices are forced to transmit large volumes of compressed data that may prove irrelevant to the cloud's current queries, wasting precious communication resources. The challenge manifests in three critical dimensions.

First, the inherent mismatch between how data is transmitted from edge devices and how the data of interest is queried at the cloud causes \emph{data drift}~\cite{shubha2023adainf,mallick2022matchmaker}. In edge-cloud systems where bandwidth is extremely limited, transmission efficiency is key. Existing works~\cite{yao2020deep,hu2020starfish,hojjat2024limitnet} cannot adapt to dynamically changing query requirements, forcing systems to either transmit all compressed data regardless of relevance or repeatedly retransmit data when queries change.

\begin{figure}[tb]
\large
\centerline{\includegraphics[width=0.85\linewidth]{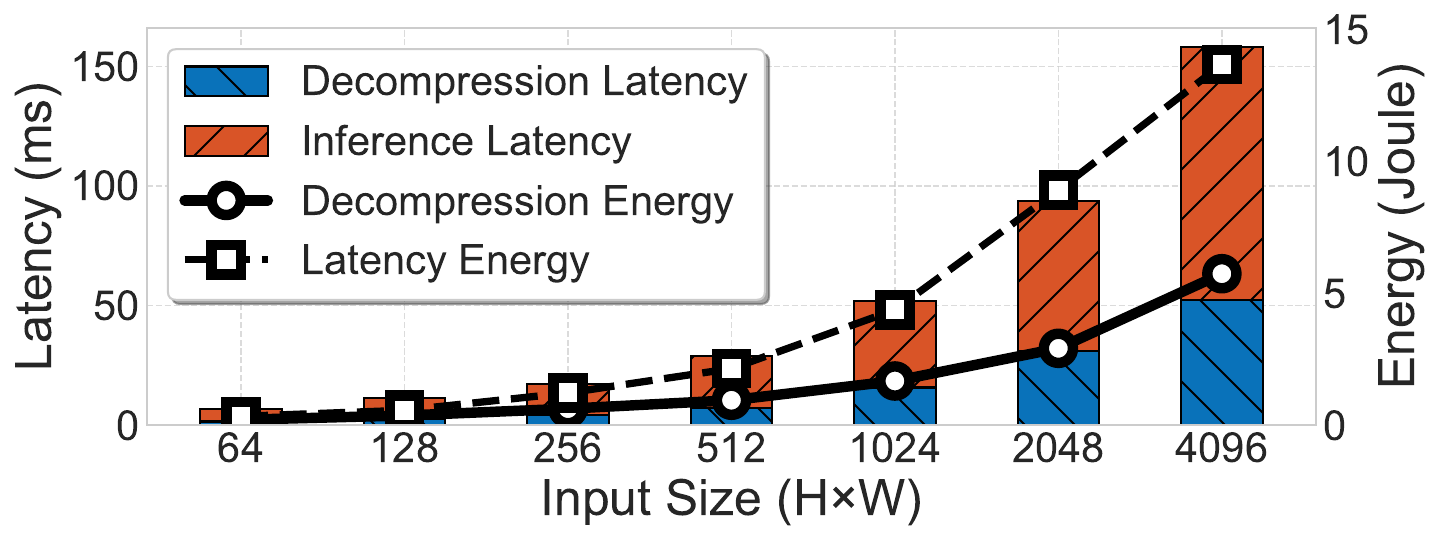}}
\vspace{-0.5cm}
\caption{Comparison of latency and energy consumption when performing YOLOv8 inference with different input sizes on NVIDIA Jetson Orin Nano.}
\Description{Comparison of latency and energy consumption when performing YOLOv8 inference with different input sizes.}
\vspace{-0.5cm}
\label{fig:resolution}
\end{figure}

Second, the overhead of \emph{decompress-infer-recompress (DIR)} at the edge creates a significant bottleneck~\cite{du2025earth+,zhang2024cosmic,furutanpey2025fool}, which holds for both static imagery and videos. As shown in Figure~\ref{fig:observation}, the analysis of the widely used DOTA-v1.0 (satellite imagery) and VisDrone 2019 (aerial imagery) datasets reveals that the regions of interest (RoI) occupies merely 12$\sim$17\% of the area in over 90\% of frames. Figure~\ref{fig:resolution} further reveals that on off-the-shelf edge hardware platforms like the NVIDIA Jetson Orin Nano, full decompression of high-resolution images consumes up to 35\% of the total processing time and 42\% of energy consumption. Even if combined with lightweight inference, that cost remains. Figure~\ref{fig:resolution} does not even report recompression costs, which may even be higher than decompression and would add on top.

For videos, DIR includes video transcoding, shown to be computation-intensive~\cite{bukhari2023video,dogga2019edge}, especially at the edge. For videos, recompression is the bottleneck: video encoding dominates end-to-end latency~\cite{du2022accmpeg}. This process would also be CPU-hungry~\cite{bukhari2023video}, consuming 60-90\% of the CPU resource and adding 5$\sim$10\% battery drain~\cite{dogga2019edge}. Consequently, in environments where connections are intermittent and video streaming is impractical, this sparsity coupled with the high cost of pixel-domain processing would lead to a massive waste of computational resources and bandwidth.

Third, the limitations of existing edge-cloud systems~\cite{li2021appealnet,banitalebi2021auto,hu2020starfish} are further exacerbated by the \emph{resource constraints} of satellite edge devices, where continuous operation is crucial, yet battery life and energy budgets are limited~\cite{li2024battery}. Existing work~\cite{yao2020deep,hojjat2024limitnet,huang2022real,laskaridis2020spinn}  fully decompress or upload compressed data to the cloud, without adapting to dynamically changing queries. For instance, DeepCOD~\cite{yao2020deep} processes all compressed features without dynamic adaptation. LimitNet~\cite{hojjat2024limitnet} requires uploading all compressed data to the cloud, resulting in significant energy consumption.
Although conserving energy is not a primary objective of existing edge computing literature, it becomes so when edge computing is deployed on a resource-constrained device such as a satellite built with off-the-shelf hardware~\cite{orbitalEdge,denby2019orbital,berck2025launch,yaacoub2025fault,narayana2020hummingbird}.

\fakepar{Contribution} To address these challenges, we design DynaFilter, a cloud-driven dynamic filtering technique. 
DynaFilter targets extremely bandwidth-constrained environments where data is represented by either static images or video streaming is applied for limited time intervals, for instance, whenever a reliable connection to the ground is available~\cite{satnogs}. 
It enables filtering data that is relevant to a cloud's query before transmission, enabling analysis in a highly-constrained communication setting.

Unlike prior work built on a paradigm where the edge uploads data regardless of the specific application interests, while the cloud analyzes it, DynaFilter fosters a different execution model. 
The cloud dynamically configures filtering parameters based on current queries representing application interests. 
The edge executes compressed-domain filtering to identify potential RoIs without full decompression.

Our key insight enabling this functionality is that low-level JPEG/Video features, such asDC coefficients/AC energy, and motion vectors in the compressed-domain exhibit strong correlations with high-level semantics, for instance, object categories in a classificaiton task. Backed by an experimental foundation described next, we establish a mapping between compressed data and dynamic filtering, enabling accurate RoI prediction with minimal computational overhead. Extensive evaluations show that DynaFilter reduces decompressed data size by $1.6\times$$\sim$$7.1\times$, compared to state-of-the-art baselines. Furthermore, it decreases energy consumption by 43.1$\sim$88.6\% and achieves $1.6\times$$\sim$$3.0\times$ speedup in inference latency.

Overall, we make the following contributions:
\begin{enumerate}
\item
We reveal the data drift challenge in satellite edge computing, where compressed data stored at the edge does not align with dynamically changing queries at the cloud, representing application interests.
\item 
We propose a new cloud-driven dynamic filtering technique, enabling the cloud to dynamically configure filtering parameters for edge devices and eliminating the need for full inference or recompression.
\item 
We design a compressed-domain feature mapper that establishes a precise mapping between high-level semantics and low-level compressed-domain features.
\item 
We develop a multimodal feature fusion filter that enhances filtering accuracy by incorporating temporal-spatial features for dynamic video scenarios.
\end{enumerate}

In the rest of the paper, Section~\ref{background} provides background information on compressed-domain features. Section~\ref{motivation} motivates the work, whereas Section~\ref{Overview} describes the system overview. Section~\ref{Design} details the design. Section~\ref{Implementation} presents the implementation, and Section~\ref{Evaluation} reports experimental results. Section~\ref{Related} presents related work, while Section~\ref{Discussion} discusses limitations and extensions. Section~\ref{Conclusion} ends the paper.

\section{Background}\label{background}
We design DynaFilter to leverage the inherent organization of discrete cosine transform (DCT) coefficients~\cite{chang1995manipulation} within the JPEG standard~\cite{wallace1991jpeg}, which systematically partitions image information into frequency-domain components~\cite{skodras2002jpeg}. 
This design principle enables DynaFilter to perform semantic analysis directly in the compressed domain. 

\fakepar{Discrete cosine transform (DCT)}
JPEG divides images into 8×8 pixel blocks and applies DCT to each block:
\begin{equation}
\footnotesize
F(u,v) = C(u)C(V)\sum_{x=0}^{7} \sum_{y=0}^{7}f(x,y)\cos[\frac{(2x+1)u\pi}{16}]\cos[\frac{(2y+1)v\pi}{16}]
\label{DCT}
\end{equation}
where $f(x,y)$ represents the pixel value at spatial position $(x,y)$, $F(u,v)$ is the DCT coefficient at frequency position, $C(u)$ and $C(v)$ are normalization factors.

\fakepar{Discrete cosine (DC) coefficient}
The DC coefficient, located at position $(0,0)$ in the DCT matrix, represents the average brightness of the 8×8 block and is calculated as:
\begin{equation}
DC_{coefficient} = F(0,0) = \frac{1}{4}\sum_{x=0}^{7}\sum_{y=0}^{7}f(x,y)
\label{DC_Coefficient}
\end{equation}

The DC coefficient~\cite{qin2022jpeg} difference between neighboring blocks is often encoded rather than the absolute value, leveraging this spatial correlation for additional compression.

\fakepar{Alternating current (AC) energy}
 The remaining 63 coefficients (where $u>0$ or $v>0$) are AC coefficients~\cite{yu2024reversible}, representing higher frequency details within the block (e.g., edges, textures, and fine-grained features). The AC energy is defined as the sum of absolute values of all AC coefficients:
\begin{equation}
AC_{energy} = \sum_{u=0}^{7}\sum_{v=0}^{7}|F(u,v)| - |F(0,0)|
\label{AC_Energy}
\end{equation}

The AC energy effectively measures the texture richness of an image. Regions with high AC energy typically contain edges, textures, or object boundaries, while background regions have low AC energy.

\section{Motivation}\label{motivation}
Our work is motivated by the challenges in satellite edge computing, where orbiting devices must analyze high-resolution imagery and video streams under severe bandwidth on top of possibly intermittent communication channels, and subject to energy constraints. By analyzing the characteristics of remote sensing data and the processing overhead on edge devices, we derive four key observations.

\begin{figure}[tb]
\large
\centerline{\includegraphics[width=0.95\linewidth]{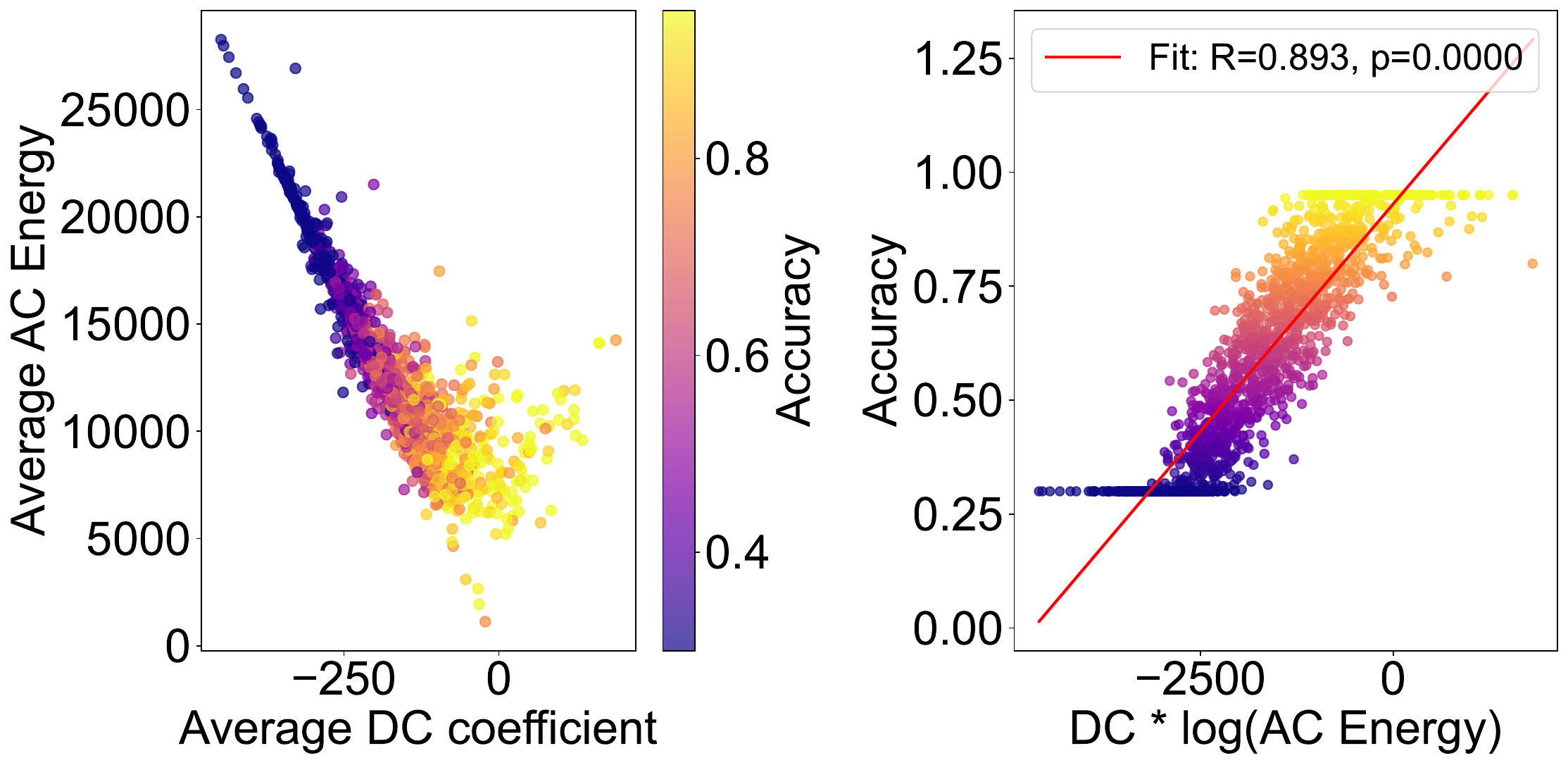}}
\vspace{-0.3cm}
\caption{Relationship between JPEG features, that is, DC coefficient and AC energy, and inference accuracy.}
\Description{Relationship between JPEG features (i.e., DC coefficient and AC energy) and inference accuracy.}
\vspace{-0.3cm}
\label{fig:observation1}
\end{figure}

\begin{figure}
\vspace{-0.3cm}
\setlength{\abovecaptionskip}{0pt}
\setlength{\belowcaptionskip}{0pt}
\centering
\subfigure[DOTA-v1.0 Dataset]{
\begin{minipage}[b]{0.485\linewidth}
\includegraphics[width=1\linewidth]{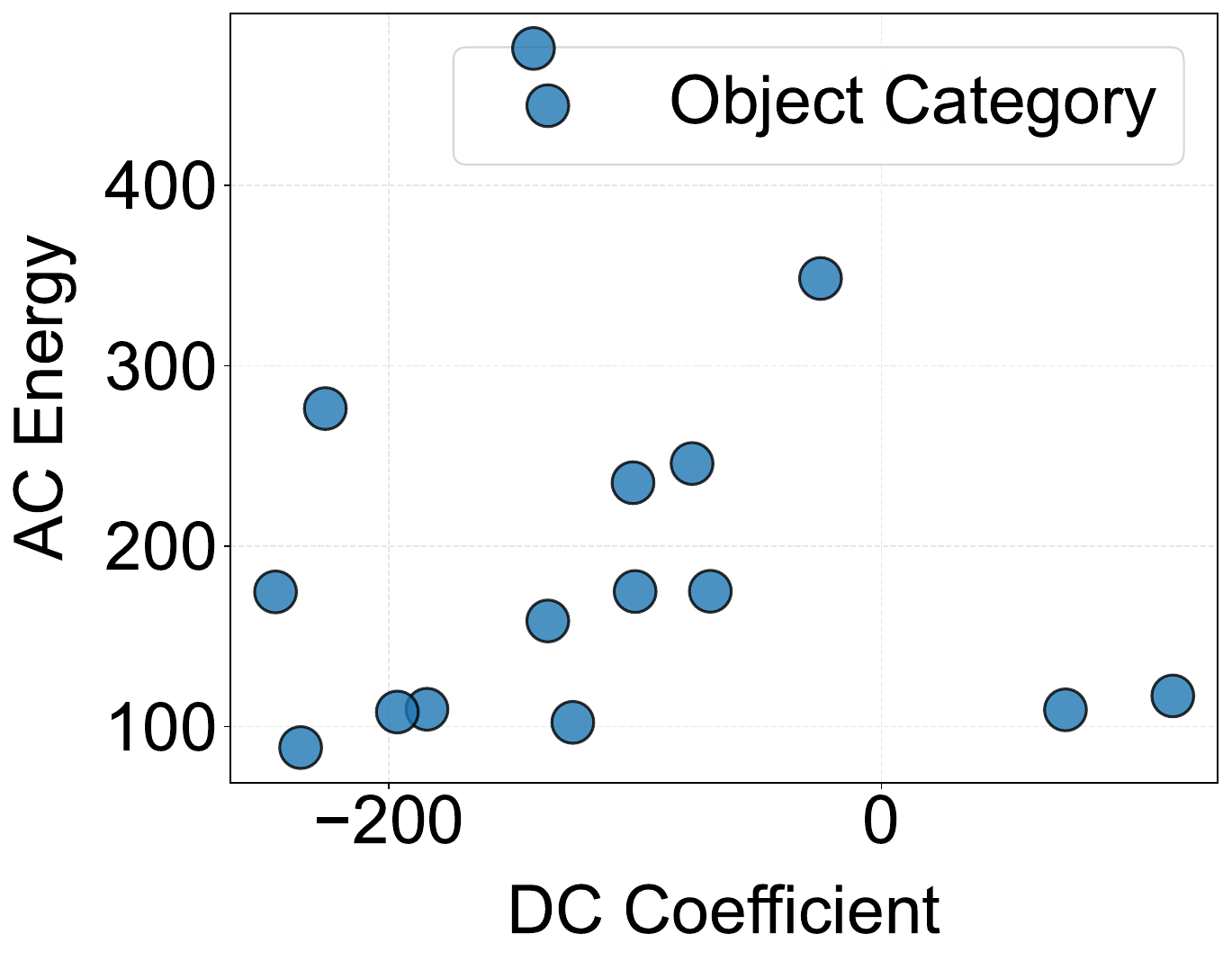}
\end{minipage}}
\subfigure[VisDrone 2019 Dataset]{
\begin{minipage}[b]{0.485\linewidth}
\includegraphics[width=1\linewidth]{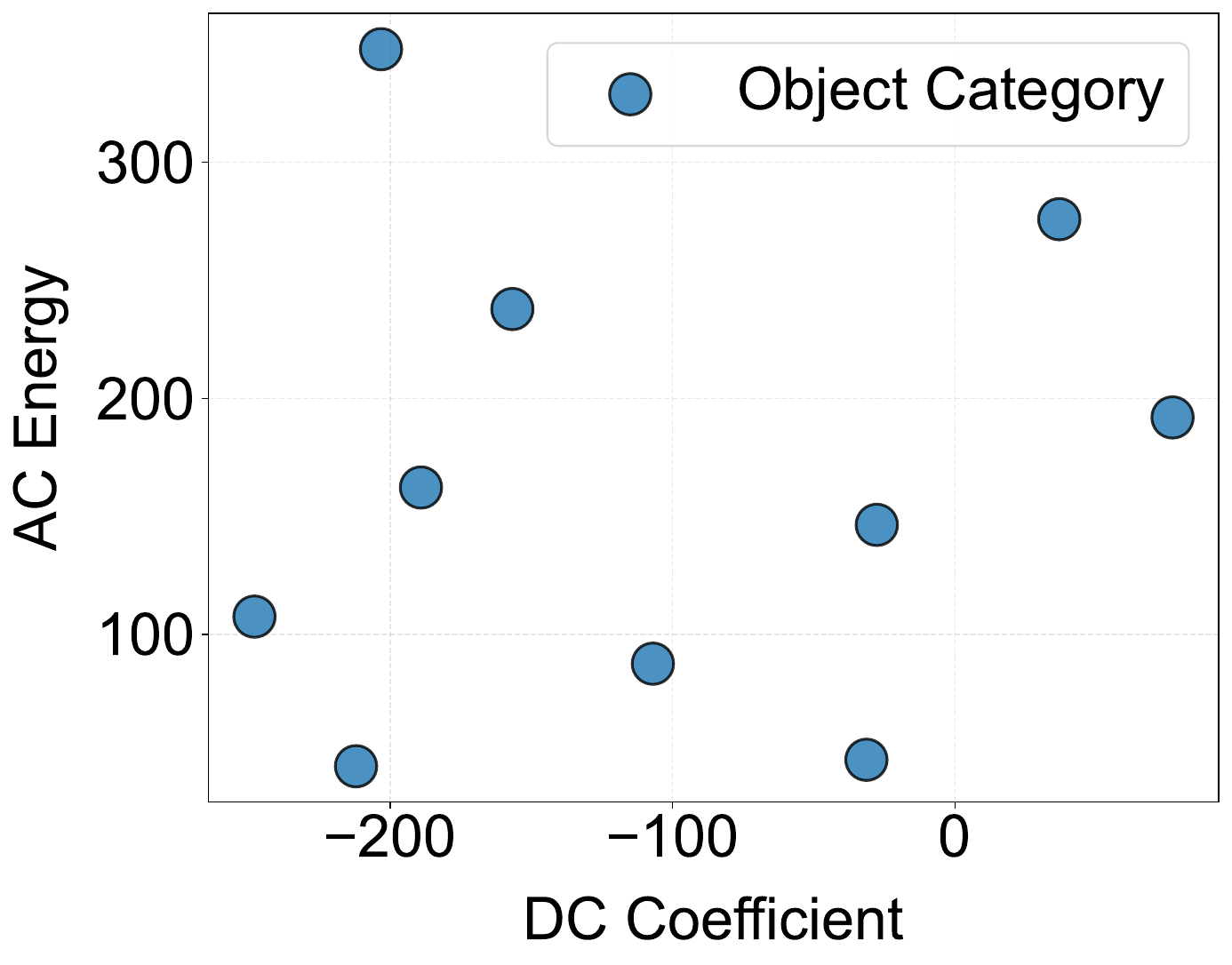}
\end{minipage}}
\caption{Distribution of JPEG features, that is, DC coefficient and AC energy, for each category.}
\Description{Distribution of JPEG features for each category.}
\vspace{-0.3cm}
\label{fig:observation2}
\end{figure}

First, we find a strong relationship between low-level JPEG features, that is, DC coefficients and AC energy, and high-level semantics, for example,  inference accuracy, exist, as shown in Figure~\ref{fig:observation1}.
Accuracy here refers to the inference mAP, that is, mean average precision, achieved when the DNN model is run only on regions matching specific AC/DC signatures.
This correlation is consistent across different object categories in Figure~\ref{fig:observation2}, where the JPEG features for each category have distinct distributions. 
Note that the DC coefficients for object regions tend to cluster within specific ranges depending on the object category, creating identifiable patterns in the compressed-domain.
\begin{tcolorbox}[myobservation=1]
Low-level compressed-domain features are closely related to high-level semantics.
\end{tcolorbox}

% \begin{figure}
% \vspace{-0.3cm}
% \setlength{\abovecaptionskip}{0pt}
% \setlength{\belowcaptionskip}{0pt}
% \centering
% \subfigure[Relationship between DC coefficient /AC energy and accuracy]{
% \begin{minipage}[b]{0.5\linewidth}
% \includegraphics[width=1\linewidth]{Figures/DC_coefficient_AC_energy_accuracy.pdf}
% \end{minipage}}
% \subfigure[Relationship between combined JPEG features and accuracy]{
% \begin{minipage}[b]{0.47\linewidth}
% \includegraphics[width=1\linewidth]{Figures/combined_JPEG_features_accuracy.pdf}
% \end{minipage}}
% \caption{Relationship between JPEG features (i.e., DC coefficient and AC energy) and inference accuracy.}
% \Description{Relationship between JPEG features (i.e., DC coefficient and AC energy) and inference accuracy.}
% \vspace{-0.3cm}
% \label{fig:observation1}
% \end{figure}

\begin{figure}[tb]
\vspace{-0.2cm}
\large
\centerline{\includegraphics[width=\linewidth]{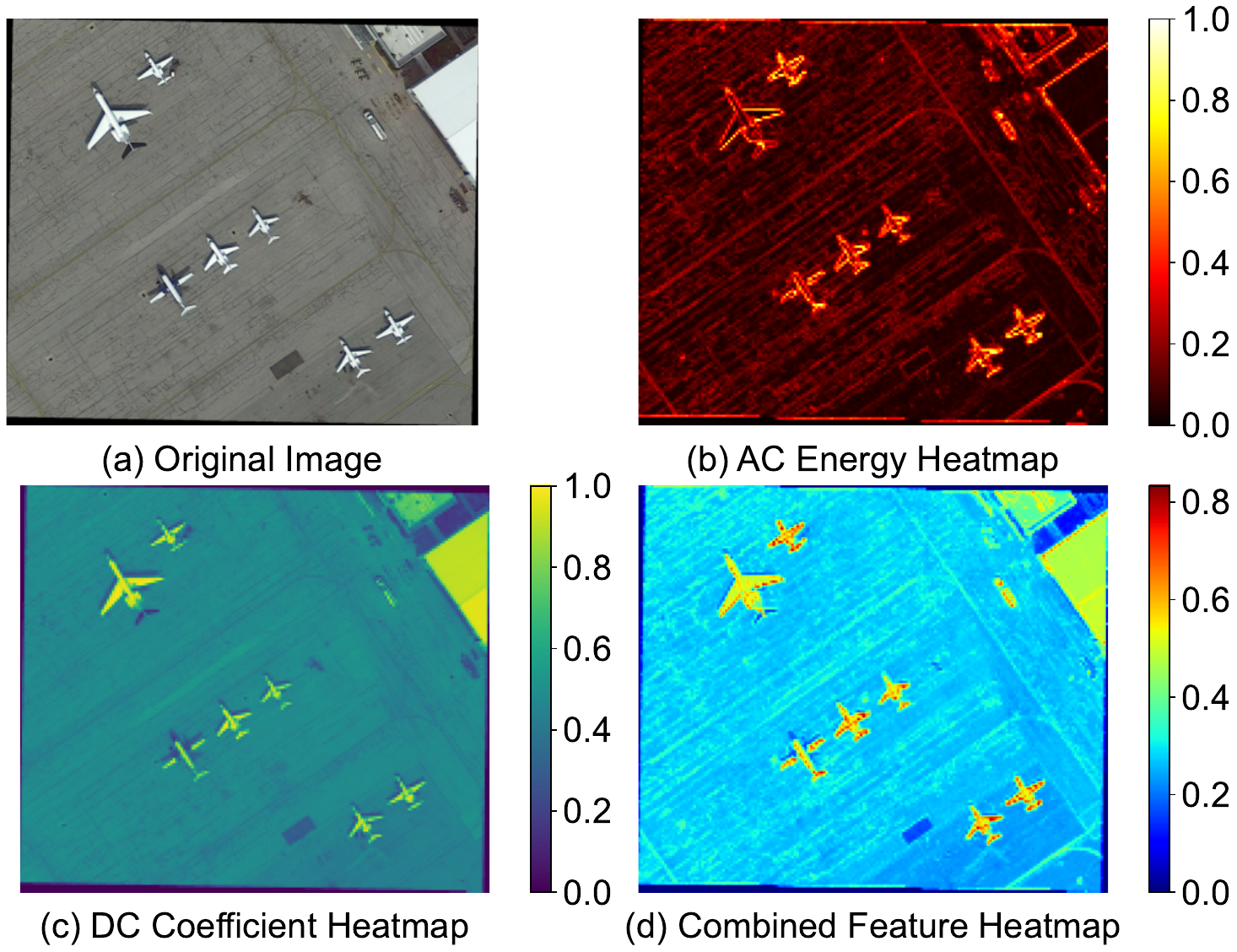}}
\vspace{-0.3cm}
\caption{JPEG features for the plane category.}
\vspace{-0.5cm}
\Description{JPEG features for the plane category.}
%\vspace{-2mm}
\label{fig:observation3}
\end{figure}

As shown in Figure~\ref{fig:observation3}(a) as an example, the original image contains multiple RoIs. 
Figure~\ref{fig:observation3}(b) shows that object regions typically exhibit AC energy values 15$\sim$25$\times$ higher than background regions, creating a clear distinguishable boundary in the compressed-domain. Similarly, Figure~\ref{fig:observation3}(c) shows that DC coefficients provide complementary information about illumination conditions and large-scale structures.
To leverage both features, we combine them as $DC\times \log(AC \ Energy)$. Figure~\ref{fig:observation3}(d) shows a highly accurate representation of potential RoI by preserving the high-frequency detail sensitivity of AC energy, while incorporating the illumination and structural information from DC coefficients.

\begin{tcolorbox}[myobservation=2]
Low-level JPEG compressed-domain features enable accurate RoI-based inference.
\end{tcolorbox}

\begin{figure}
\vspace{0pt}
\vspace{-0.2cm}
\setlength{\abovecaptionskip}{0pt}
\setlength{\belowcaptionskip}{0pt}
\centering
\subfigure[Inference latency]{
\begin{minipage}[b]{0.485\linewidth}
\includegraphics[width=1\linewidth]{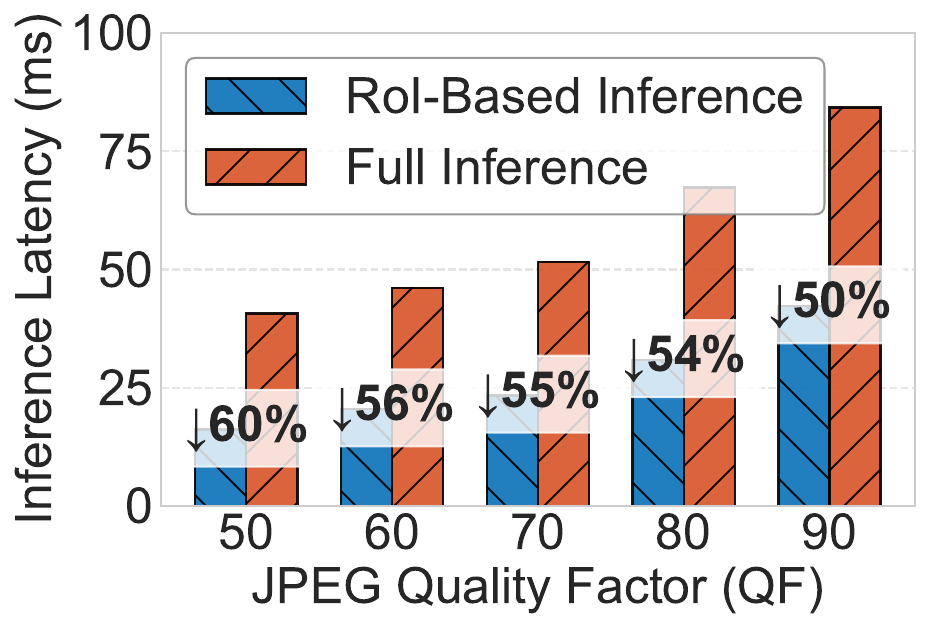}
\end{minipage}}
\subfigure[Inference accuracy]{
\begin{minipage}[b]{0.485\linewidth}
\includegraphics[width=1\linewidth]{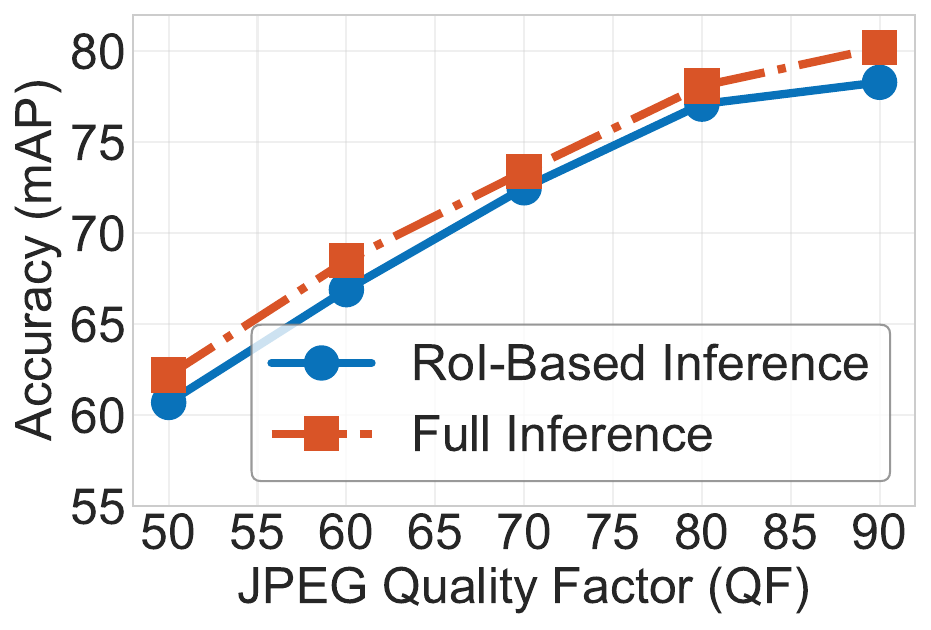}
\end{minipage}}
\caption{Comparison of RoI-based Inference and Full Inference with different JPEG quality factors.}
\Description{Comparison of RoI-based Inference and Full Inference with different JPEG quality factors.}
\vspace{-0.5cm}
\label{fig:observation4}
\end{figure}

To confirm this intuition, we compare full inference against RoI-based inference. As shown in Figure~\ref{fig:observation4}(a), RoI-based inference achieves a $2.0\times$$\sim$$2.5\times$ speedup in inference latency across different JPEG quality factors. Figure~\ref{fig:observation4}(b) confirms that this is not detrimental to accuracy, showing a less than 2\% loss. This demonstrates that processing only the regions identified by compressed-domain analysis is a viable strategy.
Note that DynaFilter aims to spare the prohibitive overhead of the DIR pipeline rather than just skipping background blocks. Even in scenarios where RoIs occupy a large area, DynaFilter remains effective by selectively decoding only the relevant blocks as determined by the cloud-driven filters, thus avoiding the recompression overhead.

\begin{tcolorbox}[myobservation=3]
RoI-based inference can achieve high energy efficiency without compromising accuracy.
\end{tcolorbox}

\begin{figure}
\vspace{0pt}
\vspace{-0.2cm}
\setlength{\abovecaptionskip}{0pt}
\setlength{\belowcaptionskip}{0pt}
\centering
\subfigure[Original frame]{
\begin{minipage}[b]{0.4625\linewidth}
\includegraphics[width=1\linewidth]{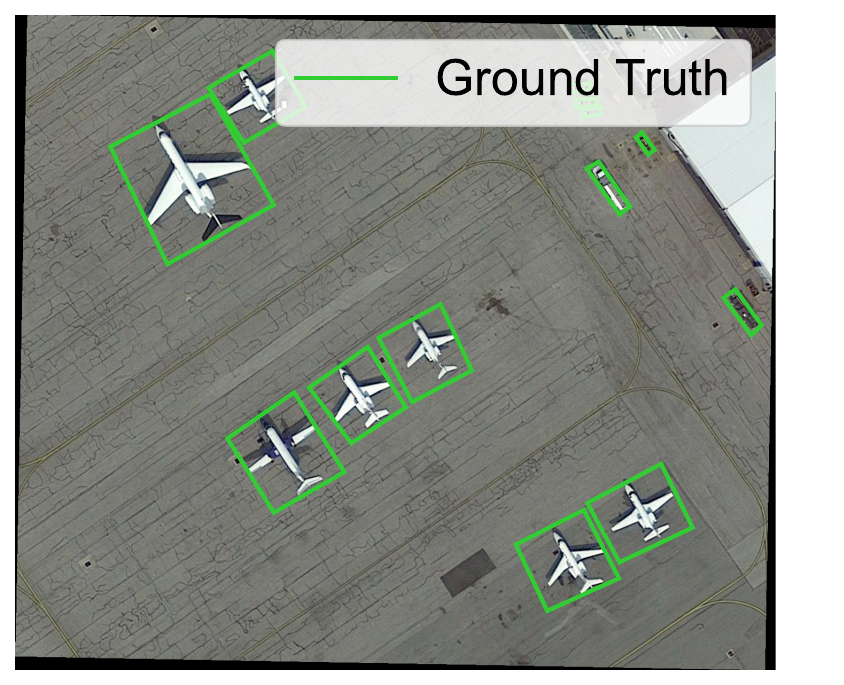}
\end{minipage}}
\subfigure[Motion intensity]{
\begin{minipage}[b]{0.5075\linewidth}
\includegraphics[width=1\linewidth]{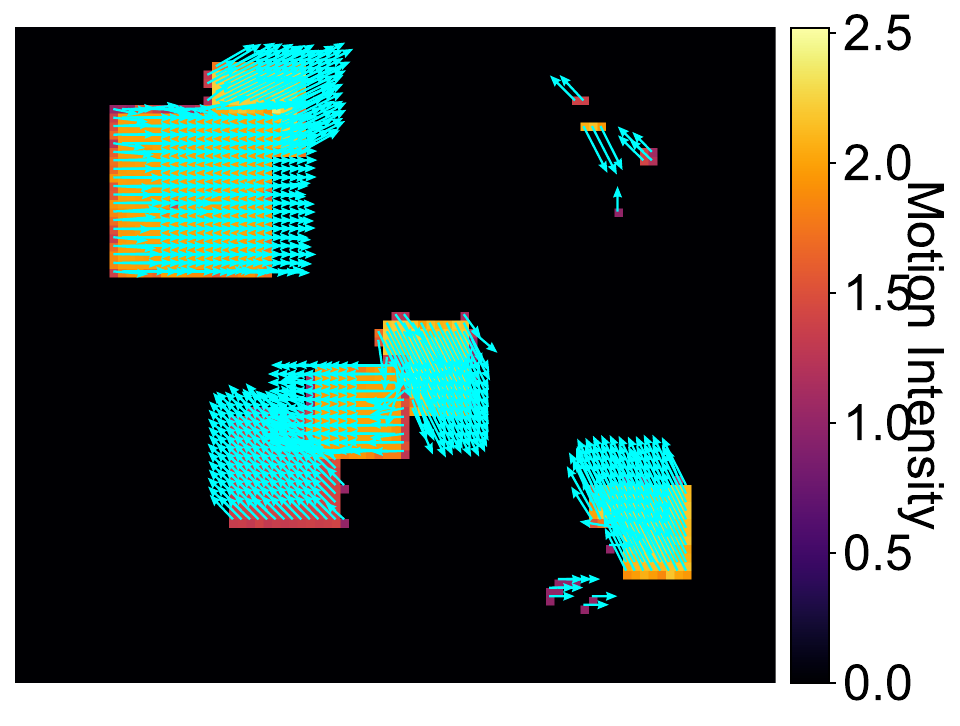}
\end{minipage}}
\caption{VisDrone 2019 dataset: Spatial correlation between ground objects and compressed-domain MVs.}
\Description{Spatial correlation analysis between ground truth objects and compressed domain motion vectors on the VisDrone 2019 dataset.}
\vspace{-0.5cm}
\label{fig:observation5}
\end{figure}

For satellite video monitoring, traditional pixel-domain methods, for example, optical flow techniques~\cite{ilg2017flownet,jin2025flovd}, are computationally prohibitive for edge devices. However, we observe that video codecs, such as H.264/HEVC, inherently encode temporal dynamics as motion vectors during compression. To study this aspect, we extracted motion vectors from compressed video sequences in the VisDrone 2019 dataset. As shown in Figure~\ref{fig:observation5}, the magnitude of motion vectors exhibits a strong spatial correlation with moving targets, separating them apart from the static background. This implies that motion can be identified directly from the compressed bitstream, enabling DynaFilter to locate dynamic RoIs with negligible computational cost before full decoding.

\begin{tcolorbox}[myobservation=4]
Motion semantics in video streams are natively encoded in bitstream motion vectors.
\end{tcolorbox}

The observations we articulated provide a stepping stone for the design of DynaFilter, described next.

\begin{figure*}[tb]
\large
\centerline{\includegraphics[width=0.9\linewidth]{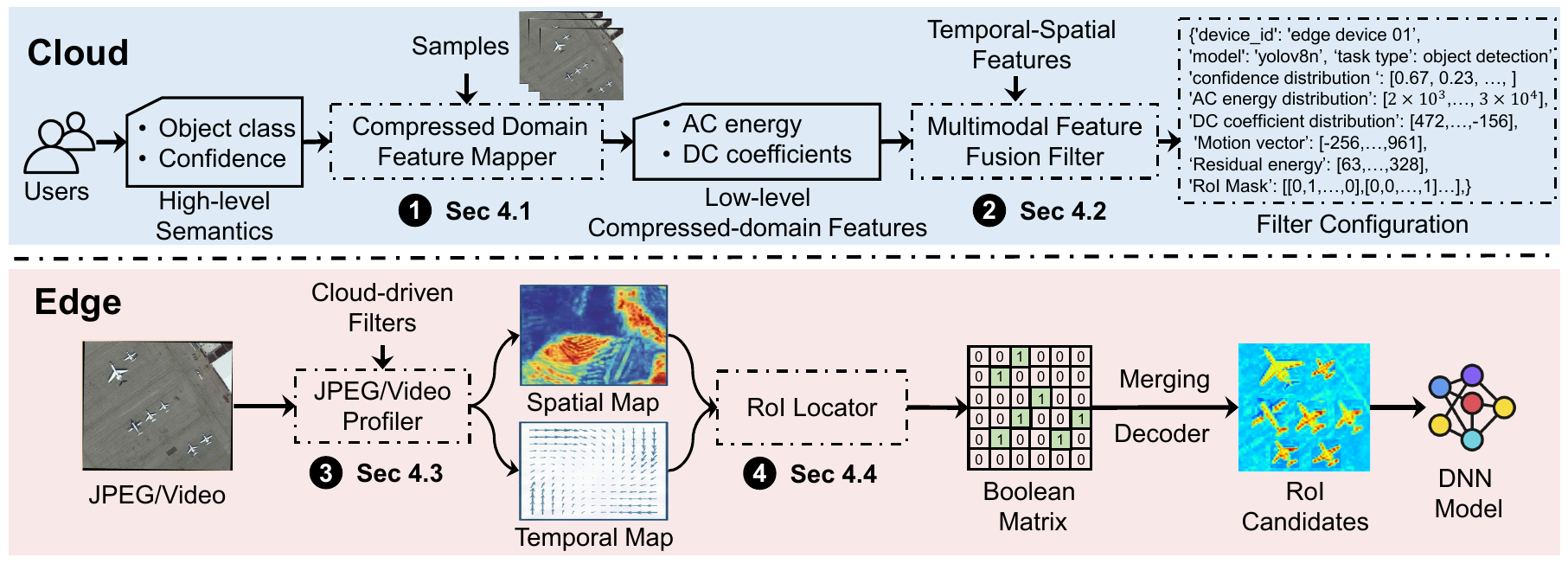}}
\caption{Two stages in DynaFilter: multimodal filter generation at the cloud for converting high-level semantics queries into compressed-domain filter configurations, and RoI-based inference at the edge for applying filters directly in the compressed-domain and performing partial decompression and inference only on potential RoIs.}
\Description{DynaFilter consists of two stages: multimodal filter generation at the cloud for converting high-level semantics queries into compressed-domain filter configurations, and RoI-based inference at the edge for applying filter configurations directly in the compressed-domain and performing partial decompression and inference only on potential RoIs.}
\vspace{-0.3cm}
\label{fig:framework}
\end{figure*}

\section{System Overview} \label{Overview}
Figure~\ref{fig:framework} shows the architecture of DynaFilter, which consists of two stages designed to bridge the semantic gap between cloud queries and compressed data at the edge.

\fakepar{Filter generation at the cloud} This stage aims to translate high-level semantic queries into multimodal compressed-domain filter configurations. 
In step \ding{182}, the compressed-domain feature mapper takes object categories and confidence thresholds as input. It utilizes a neural network to determine corresponding baseline thresholds for both spatial features, that is, AC energy and DC coefficients, and temporal features, that is, motion vectors and residual energy. 
In step \ding{183}, the multimodal feature fusion filter enhances adaptability by incorporating dynamic context. It directly analyzes motion vector statistics from historical bitstreams to generate motion-aware thresholds and refine tracking parameters, creating a robust filter configuration that adapts to both texture and motion changes. 

\fakepar{Compressed-domain filtering at the edge} This stage executes selective RoI-based inference directly on the input bitstreams. 
In step \ding{184}, the JPEG/video codec profiler parses the bitstream header to identify the modality and routes data accordingly. It extracts DCT coefficients for static content (JPEG or I-frames) to capture spatial texture, and extracts motion vectors (MVs) and residuals for dynamic video content (P/B-frames) to capture temporal motion intensity. 
In step \ding{185}, the RoI locator applies the filters configured at the cloud to extracte these features. It identifies potential RoIs using adaptive threshold matching, merges them into spatially coherent regions, and formats them for partial decompression.

\section{DynaFilter Design} \label{Design}

We detail the design of the single components in DynaFilter, according to Figure~\ref{fig:framework}.

\begin{figure}[tb]
\vspace{-0.2cm}
\large
\centerline{\includegraphics[width=.95\linewidth]{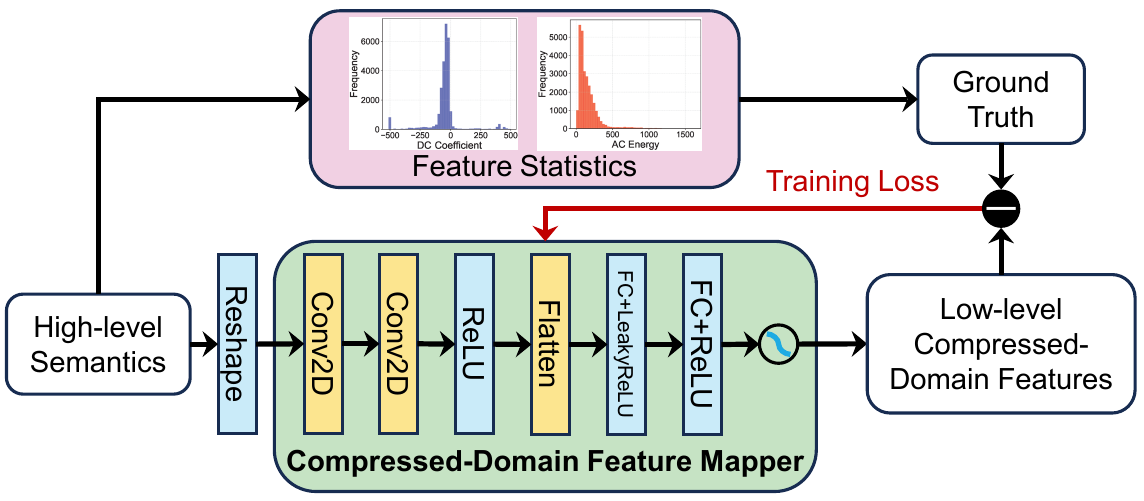}}
\caption{Compressed-domain feature mapper.}
\Description{Compressed-domain feature mapper.}
\vspace{-0.3cm}
\label{fig:mapper}
\end{figure}

\subsection{Compressed-Domain Feature Mapper}  \label{sec:Mapper}
As shown in Figure~\ref{fig:mapper}, we develop a streamlined neural network architecture to implement the mapper while maintaining accuracy and efficiency. To support the analysis of both static images and video streams, as detailed next, we extend the mapper's capability to learn cross-modal representations.

The input to the mapper is an $(N+10)$-dimensional feature vector that includes three elements:
(1) an $N$-dimensional one-hot encoding of the object categories ($N$ represents the number of object categories);
(2) a 1-di\-men\-sional confidence threshold value required by the user;
(3) 9-di\-men\-sional historical statistics. The latter includes the mean and standard deviation of the AC energy, DC coefficients, MVs, and residual energy, along with a global normalization factor.
These inputs collectively represent the semantic parameters and contextual information—covering both texture and motion priors—necessary for accurate threshold prediction.

The first stage of the mapper is an input reshaping operation that transforms the $(N+10)$-dimensional feature vector into a structured tensor. For practical implementation, we reshape this vector into a $\lceil\sqrt{N+10}\rceil \times \lceil\sqrt{N+10}\rceil \times 1$ tensor. This operation recognizes that semantic parameters possess an inherent spatial structure where related features are positioned in proximity to each other.

After reshaping, the first hidden layer is a convolutional layer with two filters of $3\times3$ size with padding of one. This layer applies spatial filters that capture local feature interactions. The second hidden layer begins with a flattening operation to integrate the spatially extracted features into a holistic representation. This flattened vector then passes through a fully connected layer with 16 neurons using LeakyReLU~\cite{xu2015empirical}. The third hidden layer consists of a fully connected layer with 8 neurons using ReLU~\cite{agarap2018deep}, extracting the most essential information needed for the final prediction.
The output consists of four neurons with linear activation, corresponding to the predicted thresholds for AC energy ($T_{ac}$), DC coefficients ($T_{dc}$), MVs ($T_{mv}$), and residual energy ($T_{res}$). Post-processing is applied to ensure these values fall within valid ranges, for example, ensuring energy is not negative.

\begin{table}[b]
\centering
\footnotesize
\caption{Prediction accuracy of mappers.}
\vspace{-0.3cm}
\label{tab:mapper_comparison}
\begin{tabular}{ccccc}
\toprule[0.75pt] 
Dataset & Approach & Accuracy & Memory & Time \\
\midrule[0.6pt]
\multirow{3}{*}{\makecell[c]{DOTA-v1.0}} & Statistical Analysis & 75.3\% & 0.5MB & 8.3ms \\
& Random Forest & 82.7\% & 4.2MB & 15.7ms \\
& \textbf{NN (Ours)} & \textbf{91.8\%} & \textbf{0.8MB} & \textbf{12.6ms} \\
\midrule[0.6pt]
\multirow{3}{*}{\makecell[c]{VisDrone 2019}} &
Statistical Analysis & 76.2\% & 0.5MB & 8.4ms \\
& Random Forest & 83.5\% & 4.3MB & 15.9ms \\
& \textbf{NN (Ours)} & \textbf{92.3\%} & \textbf{0.9MB} & \textbf{13.1ms} \\
\bottomrule[0.75pt]
\end{tabular}
\label{tbl:method}
\end{table}

We implement two alternative approaches for comparison: a statistical analysis approach and a random forest approach. Table~\ref{tbl:method} compares these approaches on the DOTA-v1.0 and VisDrone 2019 datasets. The NN-based approach achieves the highest prediction accuracy across both datasets. This stems from the NN's ability to model complex non-linear relationships between high-level semantics and low-level compression signatures, both spatial and temporal, while leveraging feature interactions through the convolutional operation. 
The NN-based approach remains lightweight, requiring only minimal memory for parameters and achieving low execution times (12.6$\sim$13.1ms), significantly outperforming the random forest approach.
Note that our NN-based ,apper learns the specific compressed-domain features for any query. For instance, a low-AC smooth surface could still be identified if that is the query semantics.

\subsection{Multimodal Feature Fusion Filter}  \label{sec:Filter}
\begin{figure}[tb]
\vspace{-0.3cm}
\large
\centerline{\includegraphics[width=.95\linewidth]{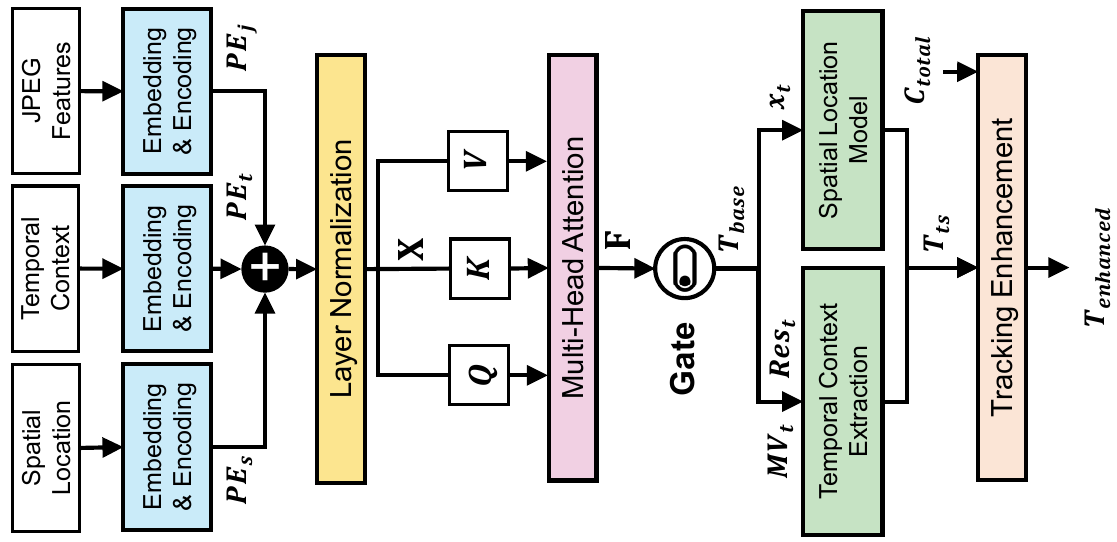}}
\vspace{-0.3cm}
\caption{Architectural overview of the multimodal feature fusion filter.}
\Description{Architectural overview of the multimodal feature fusion filter.}
\vspace{-0.5cm}
\label{fig:filter}
\end{figure}

Real-world edge intelligence applications, especially video streams, require adaptive handling of complex dynamics such as object motion, occlusion, and varying coding structures. 
We address this by integrating multiple feature modalities through a Transformer-based architecture~\cite{vaswani2017attention}. This enables DynaFilter to generate a robust, context-aware configuration that includes not only spatial thresholds ($T_{ac}, T_{dc}$) but also temporal thresholds ($T_{mv}, T_{res}$) for video analysis.
Figure~\ref{fig:filter} illustrates the filter operation, whereas Algorithm \ref{alg:fusion} details the complete process.

\begin{algorithm}[t]
\footnotesize
\caption{Multimodal Feature Fusion Process}
\label{alg:fusion}
\begin{algorithmic}[1]
\REQUIRE Semantic parameters $S$, Compressed features $C$, Motion stats $M$, Spatial features $L$
\ENSURE Enhanced filter configuration $T_{enhanced}$

\STATE $X \leftarrow \text{LayerNorm}(\text{Concat}(E_c + PE_j, E_m + PE_m, E_l + PE_l))$
\STATE $F \leftarrow \text{MultiHead}(X)$

% \STATE \COMMENT{Adaptive Threshold Generation (Spatial \& Temporal)}
\STATE $g \leftarrow \sigma(W_g \cdot F + b_g)$
\STATE $T_{ac/dc} \leftarrow \text{Compute spatial thresholds}(g, S)$
\STATE $T_{mv} \leftarrow \mu_{mv} - g_{mv} \cdot \sigma_{mv}$ 
\STATE $T_{res} \leftarrow \mu_{res} - g_{res} \cdot \sigma_{res}$ 
\STATE $T_{base} \leftarrow \{T_{ac}^{min/max}, T_{dc}^{min/max}, T_{mv}, T_{res}\}$

% \STATE \COMMENT{Temporal-Spatial Feature Enhancement}
\STATE $I_{motion} \leftarrow \frac{1}{N}\sum ||MV_t(i)||_2$ 
\STATE $x_t \leftarrow \text{Kalman filter}(x_{t-1}, z_t)$
\STATE $T_{ts} \leftarrow \text{Enhance thresholds}(T_{base}, I_{motion}, x_t)$

\IF{task is object tracking}
    % \STATE \COMMENT{Tracking Association using Motion Consistency}
    \STATE $C_{total} \leftarrow \alpha \cdot C_{space} + \beta \cdot C_{mv} + \gamma \cdot C_{reid}$
    \STATE $T_{enhanced} \leftarrow T_{ts} \cup \text{Tracking enhancement}(C_{total})$
\ELSE
    \STATE $T_{enhanced} \leftarrow T_{ts}$
\ENDIF

\RETURN $T_{enhanced}$
\end{algorithmic}
\end{algorithm}

\fakepar{Transformer-based feature encoding}
We transform heterogeneous feature inputs into a unified representation (\emph{line 1} in Algorithm \ref{alg:fusion}). To support video streams, we expand the input modalities: 
(1) compressed-domain features: historical statistics of AC energy and DC coefficients;
(2) temporal context features: historical statistics of motion vectors (MV) and residual energy derived from the video streams. It allows the model to sense the motion context (e.g., high-speed movement and stationary hovering);
(3) spatial location: the normalized coordinates of detected objects.
The unified representation $X$ is created by concatenating the encoded features from each modality with their positional encodings:
\begin{equation}
X = \text{LayerNorm}(\text{Concat}(E_j + PE_j, E_t + PE_t, E_s + PE_s)) 
\end{equation}
where $\mathbf{E}_j$, $\mathbf{E}_t$, and $\mathbf{E}_s$ represent the embedded compressed-domain features, temporal context features, and spatial location information, respectively, while $PE_j$, $PE_t$, and $PE_s$ are their corresponding positional encodings.

\fakepar{Multi-head attention fusion} 
We integrate the encoded features using multi-head attention~\cite{cordonnier2020multi} to dynamically weight the importance of different modalities (\emph{line 2} in Algorithm \ref{alg:fusion}). For instance, in high-motion video scenes, the attention mechanism automatically assigns higher weights to the temporal context features ($E_t$) to prioritize motion-based filtering over static texture matching.
The multi-head attention mechanism is formalized as:
\begin{equation}
\mathbf{F} = \text{MultiHead}(\mathbf{X}) = \text{Concat}(\text{head}_1, \dots, \text{head}_h) \mathbf{W}^O
\end{equation}
where $\text{head}_i = \text{Attention}(\mathbf{X}\mathbf{W}_i^Q, \mathbf{X}\mathbf{W}_i^K, \mathbf{X}\mathbf{W}_i^V)$.

\fakepar{Adaptive threshold generation}
We implement an adaptive gating mechanism~\cite{zhang2025e4} that generates confidence-aware thresholds (\emph{line 3} in Algorithm \ref{alg:fusion}). Crucially, we extend the gating function to output thresholds for motion vectors $T_{mv}$ and residual energy $T_{res}$ (\emph{line 4$\sim$7} in Algorithm \ref{alg:fusion}).
The gating function is formalized as:
\begin{equation}
\mathbf{g} = \sigma(\mathbf{W}_g \cdot \mathbf{F} + \mathbf{b}_g) 
\end{equation}
where $\mathbf{F}$ is the feature representation from the attention computation stage, $\mathbf{W}_g$ and $\mathbf{b}_g$ are learnable parameters, and $\sigma$ is the sigmoid function. 
The gating values then modulate the historical feature statistics to produce the final filter configuration $T_{base} = \{T_{ac}^{min}, T_{ac}^{max}, T_{dc}^{min}, T_{dc}^{max}, T_{mv}, T_{res}\}$. $T_{mv} = \mu_{mv} - g_{mv} \cdot \sigma_{mv}$ represents the minimum motion intensity required to trigger a dynamic RoI.

\fakepar{Temporal-spatial feature enhancement strategy}
This strategy aims to support object tracking tasks (\emph{line 8$\sim$10} in Algorithm~\ref{alg:fusion}). The temporal component utilizes the motion intensity ($I_{motion}$) derived directly from the bitstream:
\begin{equation}
I_{motion} = \frac{1}{N} \sum_{k=1}^{N} \sqrt{(mv_x^{(k)})^2 + (mv_y^{(k)})^2}
\end{equation}
where $mv_x$ and $mv_y$ are the motion vector components extracted by the profiler. It allows the Kalman filter-based state estimation to predict object trajectories with significantly higher accuracy, ensuring that the generated filters are spatially aligned with the predicted location of moving targets.

\fakepar{Tracking association enhancement}
For tracking tasks, we introduce a multimodal fusion cost function (\emph{line 11$\sim$13} in Algorithm~\ref{alg:fusion}) that incorporates motion consistency based on the extracted MVs:
\begin{equation}
C_{total} = \alpha \cdot C_{space} + \beta \cdot C_{mv} + \gamma \cdot C_{reid}
\end{equation}
where $C_{mv} = 1 - \cos(\vec{v}_{pred}, \vec{v}_{det})$ measures the cosine similarity between the predicted velocity vector and the detected motion vector. This direct utilization of bitstream motion data significantly reduces ID switching during occlusion compared to relying on pixel-domain inference alone. The final enhanced configuration $T_{enhanced}$ is then transmitted to the edge (\emph{line 14$\sim$16} in Algorithm~\ref{alg:fusion}).

\subsection{JPEG/Video Codecs Profiler}  \label{sec:Profiler}
To seamlessly support both static imagery (e.g., JPEG) and dynamic video streams (e.g., H.264/HEVC), our profiler adopts a unified branching architecture. It first parses the bitstream header to identify the data modality, that is, image frame or video stream, and subsequently routes the data into the corresponding feature extraction pipeline.
Crucially, our design explicitly circumvents the most computationally intensive stages of the traditional decoding loop, including inverse quantization, inverse discrete cosine transform (IDCT), and specifically for video codecs, loop filtering and motion compensation. Our profiler drastically reduces computational overhead compared to full decoding.

\subsubsection{Static Feature Extraction for JPEG Streams}
\begin{figure}[tb]
\vspace{-0.5cm}
\large
\centerline{\includegraphics[width=1\linewidth]{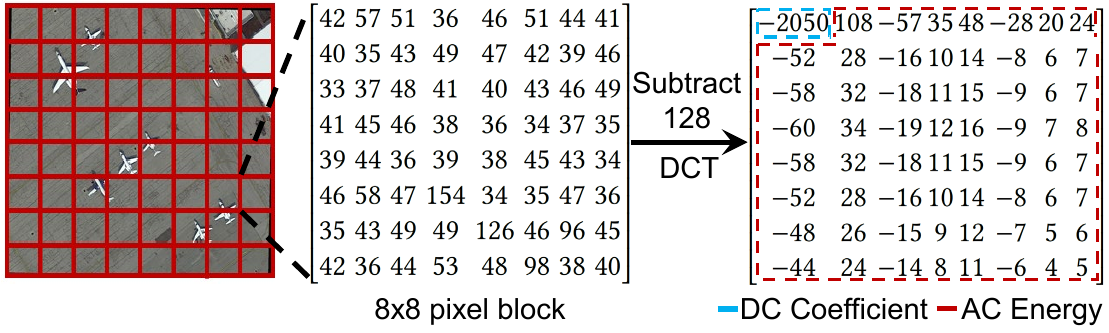}}
\vspace{-0.3cm}
\caption{JPEG profiler: it processes JPEG bitstream incrementally to extract DCT coefficients, then calculates DC coefficients and AC energy for each 8×8 block without full decompression.}
\Description{An illustration of the profiler for JPEG. It processes JPEG bitstream incrementally to extract DCT coefficients, then calculates DC coefficients and AC energy for each 8×8 block without full decompression.}
\vspace{-0.3cm}
\label{fig:proflier}
\end{figure}

The profiler uses libjpeg-turbo~\cite{libjpeg-turbo} to directly calculate the DCT coefficient matrix after entropy decoding.
As shown in Figure~\ref{fig:proflier}, the profiler processes JPEG bitstream incrementally to extract DCT coefficients, then calculates DC coefficients and AC energy for each 8×8 block without full decompression. DC coefficients capture the overall brightness of regions, which is useful for distinguishing between different object categories based on their typical luminance patterns. AC energy, on the other hand, measures the texture complexity within a block, which is indicative of whether a region contains meaningful content (such as object edges) or uniform background. These values would then be compared against the filter configuration thresholds ($T_{ac}$, $T_{dc}^{min}$, $T_{dc}^{max}$) to determine if this block is part of a potential RoI.

\subsubsection{Dynamic Feature Extraction for Video Streams}
\begin{figure}[tb]
\large
\centerline{\includegraphics[width=1\linewidth]{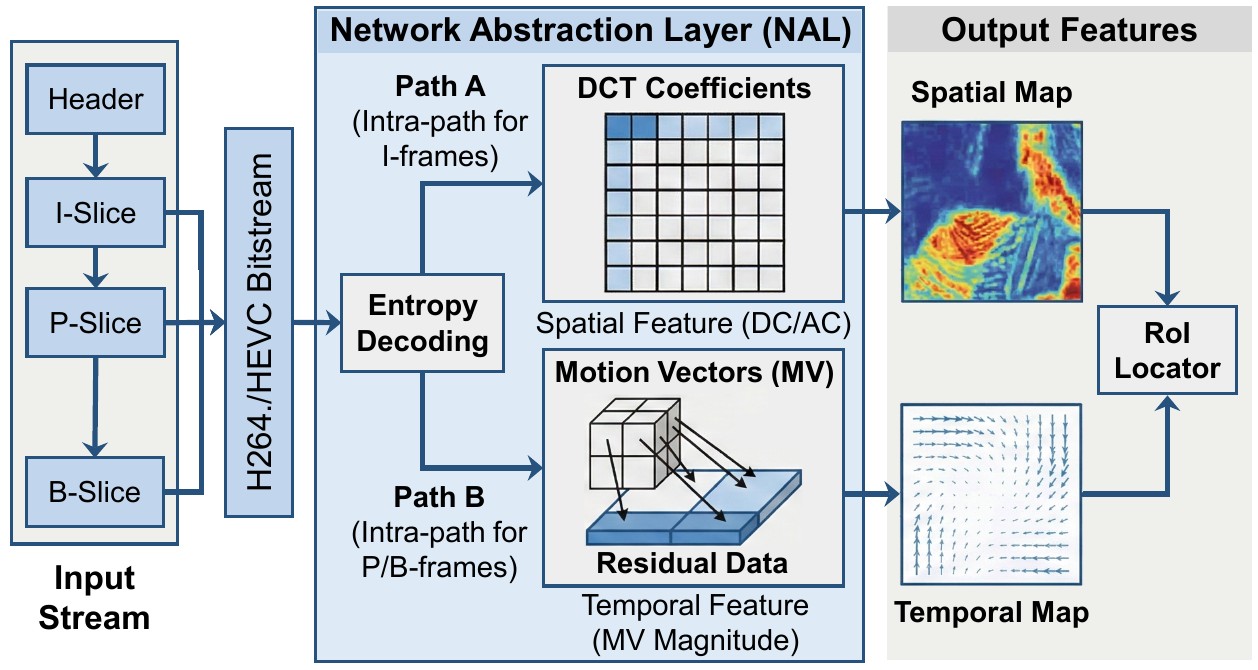}}
\vspace{-0.3cm}
\caption{Video profiler: it extracts DCT and MV features from I-Frames and P/B-Frames by parsing NAL units and entropy-decoded data, avoiding the overhead of pixel-level reconstruction and loop filtering.}
\vspace{-0.3cm}
\label{fig:proflier_video}
\end{figure}

The conventional DIR paradigm incurs prohibitive end-to-end latency. Furthermore, since resource-constrained satellite edge devices often lack dedicated hardware encoders, the recompression step consumes substantial CPU resources. Our Profiler eliminates this bottleneck by extending native support to H.264/HEVC bitstreams.
As shown in Figure~\ref{fig:proflier_video}, our profiler parses network abstraction layer (NAL) units and applies distinct feature extraction strategies based on the slice type, either keyframes Intra-coded Frame (I-Frame) or non-keyframes Predicted/Bi-predictive Frames (P/B-Frames)).

\fakepar{I-Frame processing} The processing logic for I-Frames mirrors that of JPEG, leveraging intra-prediction and DCT transforms. Our profiler extracts DCT residual coefficients and prediction modes. These features effectively capture the spatial texture distribution of video keyframes, which is used to identify RoIs in static backgrounds.

\fakepar{P/B-Frames processing} P/B-frames encode temporal changes by referencing preceding or succeeding frames. Our profiler extracts MVs and prediction residuals from the bitstream without requiring pixel-level reconstruction, by means of two separate techniques.
\begin{itemize}
    \item \textbf{Motion intensity analysis:} we calculate the per-block magnitude of the MV as $|MV| = \sqrt{mv_x^2 + mv_y^2}$. Significant MVs indicate dynamic regions within the scene, such as moving objects, providing semantic cues that static image features cannot offer.
    \item \textbf{Spatiotemporal consistency verification:} By integrating residual energy, our profiler effectively distinguishes between moving objects and illumination changes. A region exhibiting both large MV magnitude and high residual energy is identified with high probability as a dynamic RoI. We calculate the L1 norm of the quantized transform coefficients (QTC) as the residual energy using $E_{res}^{(k)} = \sum_{i} |QTC_i|$.
\end{itemize}

\subsection{RoI Locator}  \label{sec:Locator}
\begin{figure}[tb]
\large
\centerline{\includegraphics[width=1\linewidth]{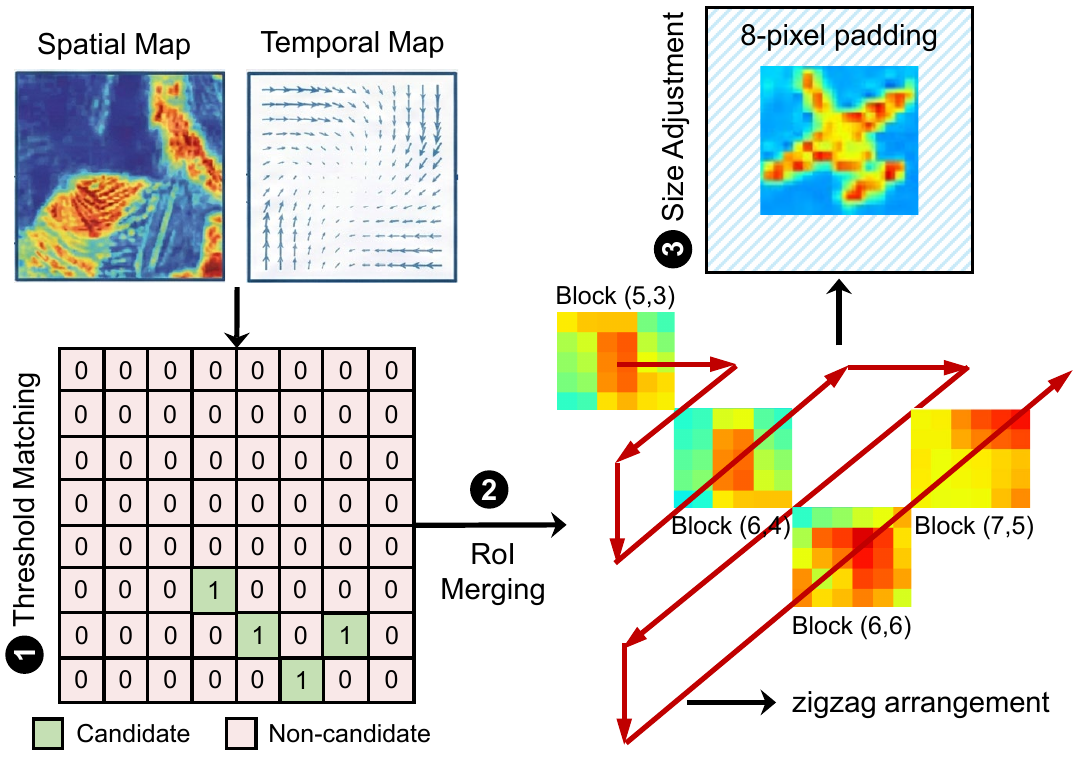}}
\vspace{-0.3cm}
\caption{The workflow of the RoI Locator processing a 32×32 pixel image (4×4 blocks with 8×8 pixels).}
\Description{Detailed workflow of the RoI Locator processing a 32×32 pixel image (4×4 blocks with 8×8 pixels).}
\label{fig:roi_locator_workflow}
\vspace{-0.3cm}
\end{figure}

The RoI Locator aims to translate the raw feature maps extracted by the profiler into actionable binary masks. As shown in Figure~\ref{fig:roi_locator_workflow}, the locator processes the feature grid through three pipelined stages: threshold matching, RoI merging, and size adjustment. For multiple queries, the cloud first calculates the union of filtering thresholds and transmits them to the edge. The  RoI Locator at the edge then extracts regions matching any of the queries, overlapping objects are subsequently separated by the classifier during the DNN inference stage. For priorities, the cloud only needs to send the filter configuration for high-priority objects.

\subsubsection{Adaptive Threshold Matching.}
The locator first generates a binary candidate map $\Theta$ by comparing block-level features against the cloud-configured filters. Figure~\ref{fig:roi_locator_workflow} (stage 1) visualizes this process for a $32 \times 32$ pixel area.

\fakepar{Spatial matching for JPEG \& I-Frames} For static content, the locator evaluates spatial texture features. In Figure~\ref{fig:roi_locator_workflow}, a block $(i, j)$ is marked as a candidate ("1") if its DC coefficient $D[i,j]$ and AC energy $A[i,j]$ fall within the specific ranges defined by the cloud (e.g., $T_{ac} \in [1500, 2000]$). It can be formalized as:
\begin{equation}
\footnotesize
\Theta_{spatial}(i, j) = \begin{cases} 
1, & \text{if } D[i,j] \in [T_{dc}^{min}, T_{dc}^{max}] \land A[i,j] \in [T_{ac}^{min}, T_{ac}^{max}] \\
0, & \text{otherwise}
\end{cases}
\end{equation}

\fakepar{Temporal matching for P/B-Frames} The matching logic shifts to motion semantics. Instead of texture, the locator evaluates the MV magnitude $|MV[i,j]|$ and residual energy $E_{res}[i,j]$. A block is flagged if it exhibits significant motion intensity or prediction error:
\begin{equation}
    \Theta_{temporal}(i, j) = \begin{cases} 
    1, & \text{if } |MV[i,j]| \ge T_{mv} \land E_{res}[i,j] \ge T_{res} \\
    0, & \text{otherwise}
    \end{cases}
\end{equation}

This ensures that we capture dynamic objects that may not have distinct static textures but strong motion signatures.

\subsubsection{RoI Merging.}
The raw binary map often contains fragmented blocks. Figure~\ref{fig:roi_locator_workflow} (stage 2) demonstrates how the RoI Locator aggregates these fragments into coherent regions. We apply connected component analysis using 4-connectivity, that is,, merging blocks sharing horizontal or vertical edges. The connectivity function $\Psi$ merges two candidate blocks $p$ and $q$ if:
\begin{equation}
\Psi(p, q) = \begin{cases} 
1, & \text{if } |p_x - q_x| + |p_y - q_y| \le 1 \\
0, & \text{otherwise}
\end{cases}
\end{equation}

After merging, the blocks are organized in a zigzag arrangement to align with standard JPEG entropy coding orders, as shown by the red arrows in Figure~\ref{fig:roi_locator_workflow}. Empirically, the zigzag pattern preserves the spatial frequency distribution of objects better than standard raster scanning, facilitating coordinate mapping for the subsequent decoding step~\cite{wallace1991jpeg,candra2017implementation}.

\subsubsection{Size Adjustment and Temporal Smoothing.}
Figure~\ref{fig:roi_locator_workflow} (stage 3) shows the padding process. If a merged region is smaller than the model's minimum receptive field (e.g., $32 \times 32$ pixels), the locator automatically expands the bounding box with padding pixels. This is crucial for detecting small objects like distant vehicles in aerial imagery.

For video content, to prevent RoI flickering caused by compression artifacts, we introduce a recursive smoothing mechanism not present in static processing:
\begin{equation}
    \Theta_{final}^{(t)}(i, j) = \alpha \cdot \Theta_{final}^{(t-1)}(i, j) + (1 - \alpha) \cdot \Theta_{current}^{(t)}(i, j)
\end{equation}
where $\Theta_{final}^{(t-1)}$ is the result from the previous frame. This ensures that the RoI Locator maintains stable tracking of moving targets even with momentary feature fluctuations.

\section{Implementation} \label{Implementation}
We implemented DynaFilter in approximately 3,800 lines of Python 3.9, along with roughly 900 lines of C++ code for low-level codec bitstream processing. We deploy the system on the widely used NVIDIA Jetson edge devices. Table~\ref{tbl:device} shows the hardware configurations of the edge devices, running Ubuntu 20.04 LTS with NVIDIA JetPack 5.1.4. 
\begin{table}[t]
\centering
\footnotesize
\caption{Hardware configuration of edge devices.}
\vspace{-2mm}
\begin{tabular}{lccccccc} \hline
\toprule[0.75pt]
\textbf{Edge GPU}  & \textbf{AI Performance}  & \textbf{DRAM}   & \textbf{Power} \\
\midrule[0.6pt]
Jetson Orin Nano  & 67TOPS (INT8)   & 8GB 102GB/s    & 7-25W   \\
Jetson AGX Orin   & 275TOPS (INT8)  & 64GB 204.8GB/s & 15-60W  \\
\bottomrule[0.75pt]
\end{tabular}
\label{tbl:device}
\vspace{-0.5cm}
\end{table}

The components at the cloud are implemented as a microservice architecture with \texttt{RESTful APIs} for interaction with edge devices. During training of the compressed domain feature mapper, the NN-based model is trained using the Adam optimizer with a learning rate of 0.001 and batch size of 64, requiring approximately 150 epochs to converge on an NVIDIA GeForce GTX 3080 GPU. The multimodal feature fusion filter is implemented as a Transformer-based encoder, including a positional encoding module and a multi-head attention mechanism. 
The filter generates a configuration in JSON format and sends it to the edge device with a payload size of only a few bytes, minimizing network overhead.

The JPEG/video codecs profiler integrates two distinct backends to support dual modalities. For static images, we utilized an extended version of libjpeg-turbo 3.0~\cite{libjpeg-turbo} to process JPEG bitstreams incrementally. We modified the library internals to expose the quantized DCT coefficients directly after the entropy decoding stage.
We integrated FFmpeg's libavcodec~\cite{tomar2006converting} for video streams with custom parsing hooks. Specifically, we intercept MVs and residuals directly from H.264/HEVC NAL units. 
% Note that the profiler retains only the calculated statistical features after feature extraction to minimize memory footprint.
To improve performance on edge devices, the RoI Locator includes SIMD optimizations, such as  NEON instructions on ARM architectures, for efficient threshold comparisons and bitmask operations. The component outputs RoI coordinates in pixel units, which are used for partial decompression and inference.

\section{Performance Evaluation} \label{Evaluation}

We describe the experimental setup first, followed by a discussion of the experimental results.

\subsection{Experiment Setup}
We train RoI-based YOLOv8~\cite{yolov8_ultralytics} for object detection using the DOTA-v1.0 dataset. For object tracking, we integrate  ByteTrack~\cite{zhang2022bytetrack} into the RoI-based YOLOv8 and train it using the VisDrone 2019 dataset.
Note that DynaFilter is model-agnostic and compatible with newer implementations, such as YOLOv11. Even with these, the fundamental trade-offs remain and thus the conclusions we draw from our experiments do not change, either.
We use the following datasets:
\begin{glossy_itemize}
    \item \textbf{DOTA-v1.0~\cite{xia2018dota,ding2021object}:} A large-scale dataset of aerial images containing 2,806 images across 15 object categories. We use this dataset for object detection. 
    \item \textbf{VisDrone 2019~\cite{zhu2021detection}:} A dataset for object detection and tracking with drones, containing 261,908 video frames across 10 object categories.
    We use it for object tracking. 
\end{glossy_itemize}
We compare DynaFilter applied to JPEG static imagery with the following baselines:
\begin{glossy_itemize}
    \item \textbf{Full Inference:} The edge device fully decompresses each frame and runs inference on the entire image.
    \item \textbf{DeepCOD~\cite{yao2020deep}:} An offloading framework that designs an encoder, trained in a data-driven manner, using compressive sensing, and then performs offloading.
    \item \textbf{LimitNet~\cite{hojjat2024limitnet}:} An efficient offloading technique; the content-aware encoder prioritizes critical data based on image content and then performs offloading.
    \item \textbf{EFilter:} A variant of DynaFilter without RoI positioning, where the edge performs compressed-domain feature extraction but processes all features.
\end{glossy_itemize}
For video streaming, instead, we consider two baselines over H.264/HEVC streams:
\begin{glossy_itemize}
\item \textbf{Full Streaming:} transmitting the complete original bitstream to the cloud for full decompression and inference.
\item \textbf{DIR:} the complete \emph{decompress-infer-recompress} pipeline we outlined in the Introduction.
\end{glossy_itemize}

In the following, we first evaluate DynaFilter performance with JPEG imagery data. Next, we deep dive into the performance with video data, and conclude by discussing figures related to system overhead.

\subsection{Accuracy, Latency, and Energy}
% For static JPEG, We first evaluate the inference latency and accuracy of different approaches on the two datasets. 
% Figure~\ref{fig:latency}(a) reports the inference latency comparison of all approaches on the two datasets.
% For DOTA-v1.0 dataset, DynaFilter achieves a $2.2\times$ speedup compared to Full Inference. The improvement is even more pronounced on VisDrone 2019 dataset ($3.0\times$ speedup). Compared to DeepCOD, LimitNet, and EFilter, DynaFilter consistently outperforms with 1.6$\sim$2.2$\times$. This efficiency stems from eliminating full decompression and focusing only on RoIs.

% As shown in Figure~\ref{fig:latency}(b), DynaFilter only loses 1.8\% on DOTA-v1.0 and 2.4\% on VisDrone 2019 respectively, compared to full inference. This minor degradation primarily stems from false negatives, where small objects fall below our compressed-domain filtering thresholds.
% DynaFilter significantly outperforms DeepCOD by 4.8\% and 5.4\% on the two datasets. This advantage arises because DeepCOD processes all compressed features without RoI filtering.
% Compared to LimitNet, DynaFilter shows consistent improves accuracy by 2.8\% and 3.8\% respectively. The limitation of LimitNet stems from the fact that it requires full decompression in the cloud, which introduces decompression artifacts and thus reduces accuracy.
% EFilter processes all compressed features indiscriminately, introducing significant noise from irrelevant background, resulting in the lowest accuracy.

\begin{figure}
\vspace{0pt}
\vspace{-0.3cm}
\setlength{\abovecaptionskip}{0pt}
\setlength{\belowcaptionskip}{0pt}
\centering
\subfigure[Inference latency]{
\begin{minipage}[b]{0.485\linewidth}
\includegraphics[width=1\linewidth]{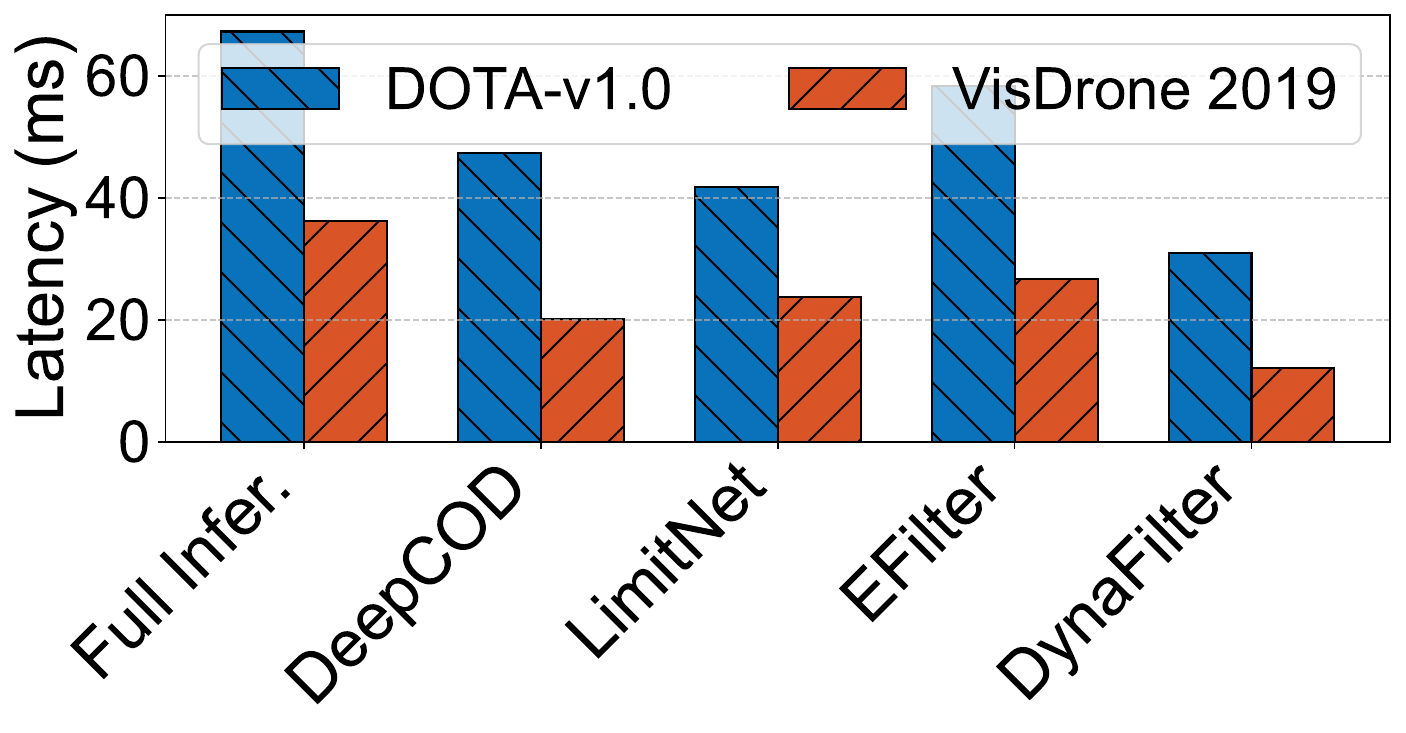}
\end{minipage}}
\subfigure[Inference accuracy]{
\begin{minipage}[b]{0.485\linewidth}
\includegraphics[width=1\linewidth]{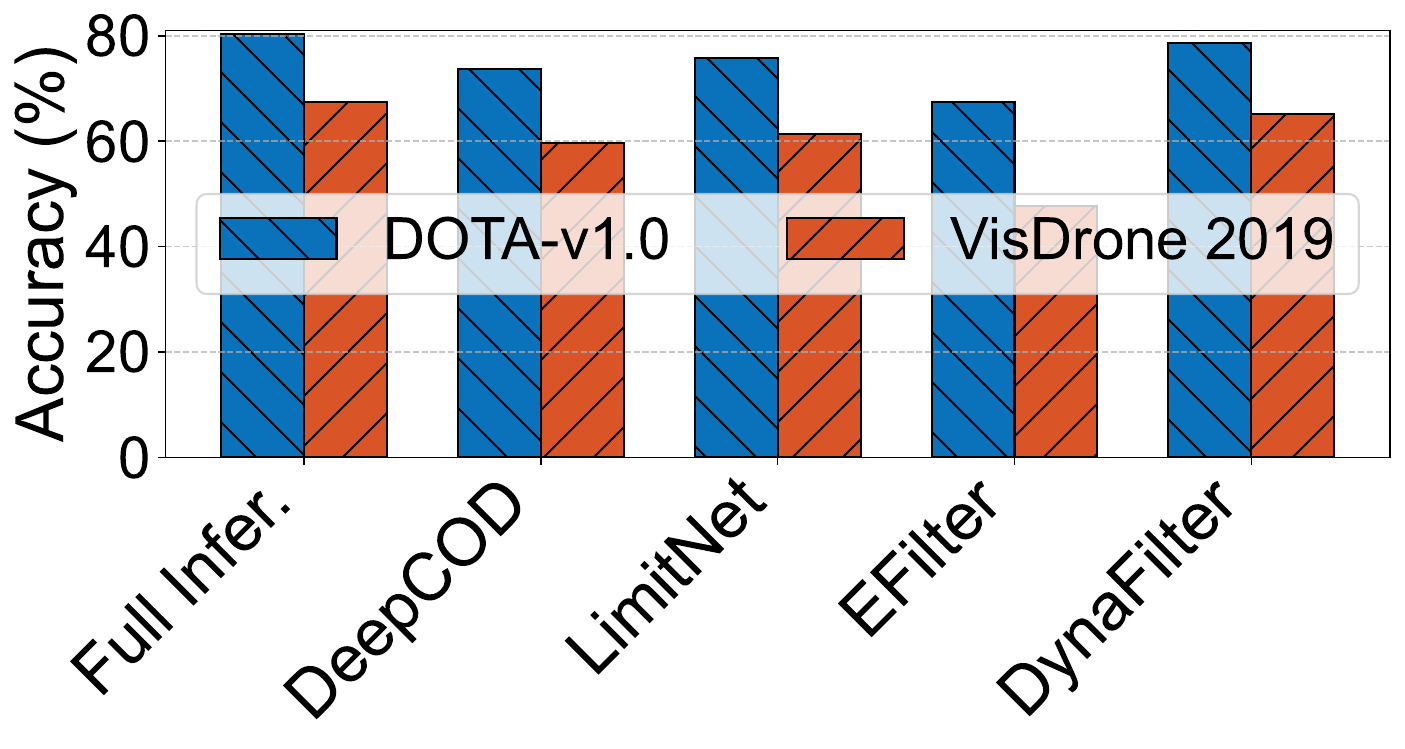}
\end{minipage}}
\caption{Inference latency and accuracy comparison on two datasets. We set the JPEG quality factor to 80.}
\Description{Inference latency and accuracy comparison on two datasets. We set the JPEG quality factor to 80.}
\vspace{-0.2cm}
\label{fig:latency}
\end{figure}

Figure~\ref{fig:latency}(a) shows that DynaFilter achieves a 2.2$\times$ and 3.0$\times$ speedup over Full Inference on DOTA-v1.0 and VisDrone 2019, respectively, and consistently outperforms DeepCOD, LimitNet, and EFilter by 1.6$\sim$2.2$\times$. This efficiency stems from eliminating full decompression and focusing solely on RoIs. 

Figure~\ref{fig:latency}(b) shows that DynaFilter maintains performance within 1.8\% on DOTA-v1.0 and 2.4\% on VisDrone of Full Inference. It outperforms DeepCOD, which lacks RoI filtering, by 5\%  and LimitNet,  which suffers from cloud-side decompression artifacts, by 3\%. EFilter performs worst due to background noise interference.

\label{Energy}
% As shown in Figure~\ref{fig:energy}, DynaFilter achieves significant energy efficiency across resolutions on NVIDIA Jetson Orin Nano. DynaFilter reduces energy by 88.6\% versus full Inference. This significant saving stems from two key factors: (1) eliminating full decompression, and (2) RoI-based selective processing.
% Compared to EFilter, DynaFilter achieves 72.1\% additional savings by selectively processing only potential RoIs. DeepCOD consumes $2.3\times$ more energy than DynaFilter due to processing all compressed features without dynamic adaptation. LimitNet uses $1.8\times$ more energy because of cloud-side decompression artifacts.

\begin{figure}[tb]
\large
\centerline{\includegraphics[width=0.6\linewidth]{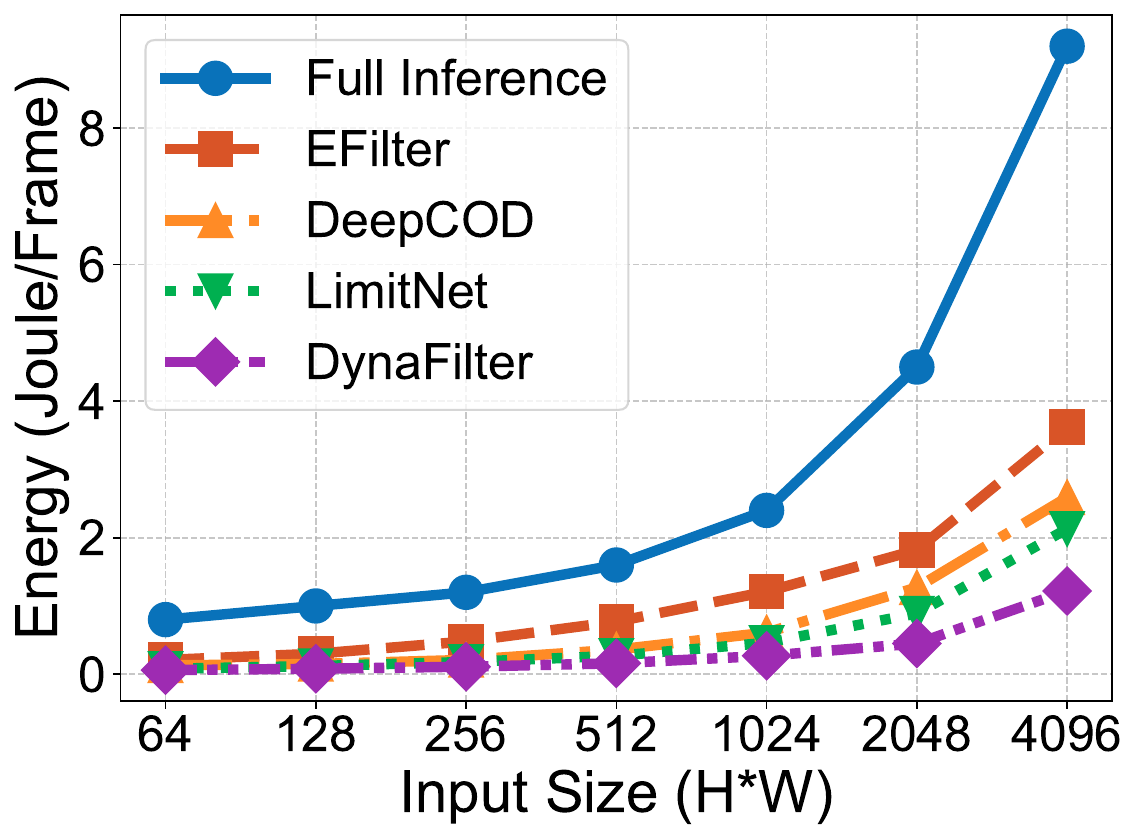}}
\vspace{-0.3cm}
\caption{Comparison of energy consumption across different input sizes.}
\Description{Comparison of energy consumption across different input sizes.}
\vspace{-0.3cm}
\label{fig:energy}
\end{figure}

Energy efficiency is critical for edge satellite computing. Figure~\ref{fig:energy} demonstrates that DynaFilter reduces energy consumption by 88.6\% compared to Full Inference. By selectively processing only potential RoIs, it achieves 72.1\% savings over EFilter. DeepCOD and LimitNet consume 2.3$\times$ and 1.8$\times$ more energy than DynaFilter, respectively, due to non-adaptive feature processing.

\subsection{Dynamic Query Filtering}
% Figure~\ref{fig:dynamic} reports a comprehensive evaluation of how different methods handle dynamically changing queries across four metrics.
% As shown in Figure~\ref{fig:dynamic}(a), DynaFilter achieves remarkably low adaptation latency across all query frequencies (i.e., the time when the cloud send a new query to the edge device to apply the new filtering parameters), achieving a $3.1\times$$\sim$$7.7\times$ improvement in adaptation speed over baselines. This advantage stems from DynaFilter's cloud-driven paradigm, which only requires transmitting compact filter configurations. In Figure~\ref{fig:dynamic}(b), DynaFilter achieves accuracy close to full inference even at high query frequencies. This stability benefits from our multimodal feature fusion that preserves semantic relevance during dynamic filtering.

% In Figure~\ref{fig:dynamic}(c), DynaFilter achieves significantly lower inference latency than the baseline across different query frequencies. This efficiency is attributed to the profiler, which incrementally processes the JPEG codestream to extract DCT coefficients.
% Additionally, the latency of DynaFilter switching between different categories in Figure~\ref{fig:dynamic}(d) is significantly lower than that of the baseline method, which demonstrates the effectiveness of DynaFilter in addressing the data drift challenge.

\begin{figure}
\vspace{-0.2cm}
\setlength{\abovecaptionskip}{0pt}
\setlength{\belowcaptionskip}{0pt}
\centering
\subfigure[Adaptation latency (ms)]{
\begin{minipage}[b]{0.485\linewidth}
\includegraphics[width=1\linewidth]{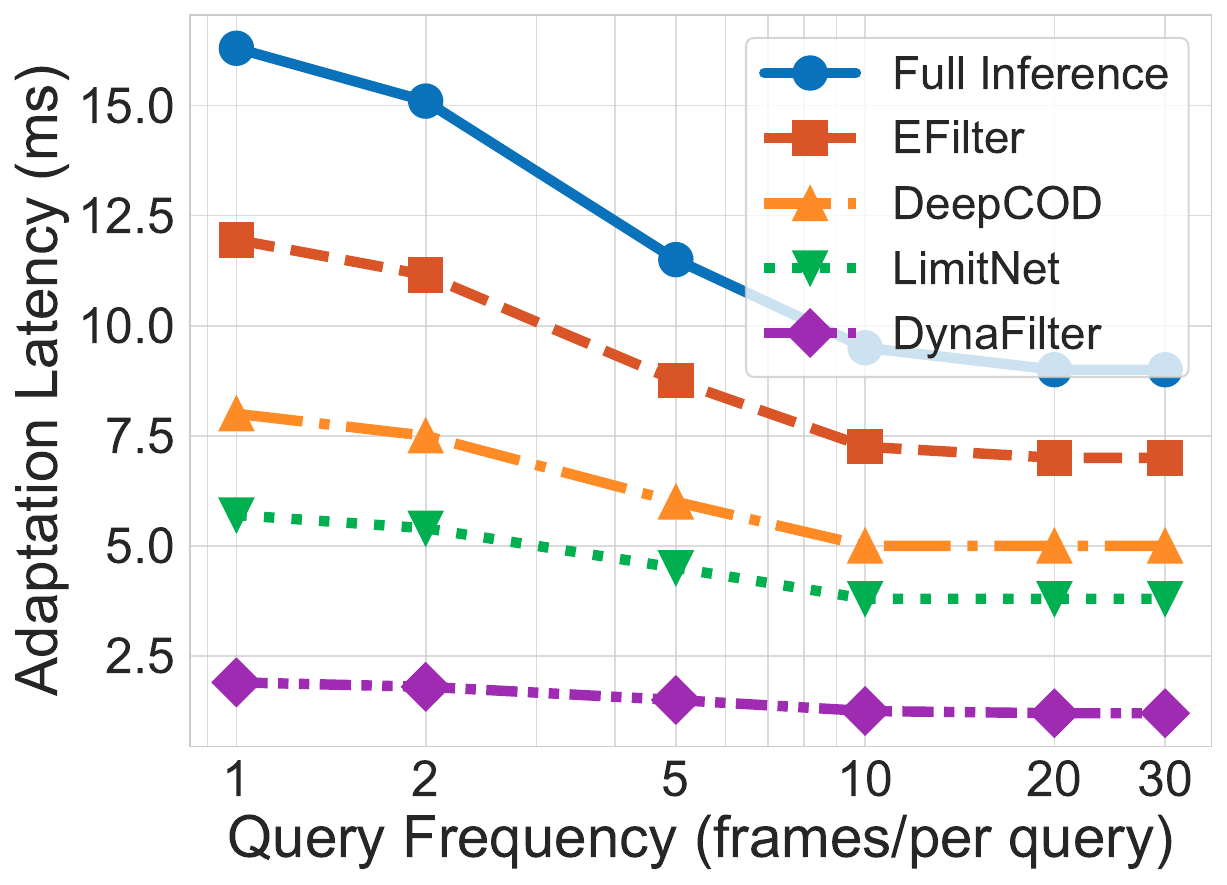}
\end{minipage}}
\subfigure[Accuracy (\%)]{
\begin{minipage}[b]{0.485\linewidth}
\includegraphics[width=1\linewidth]{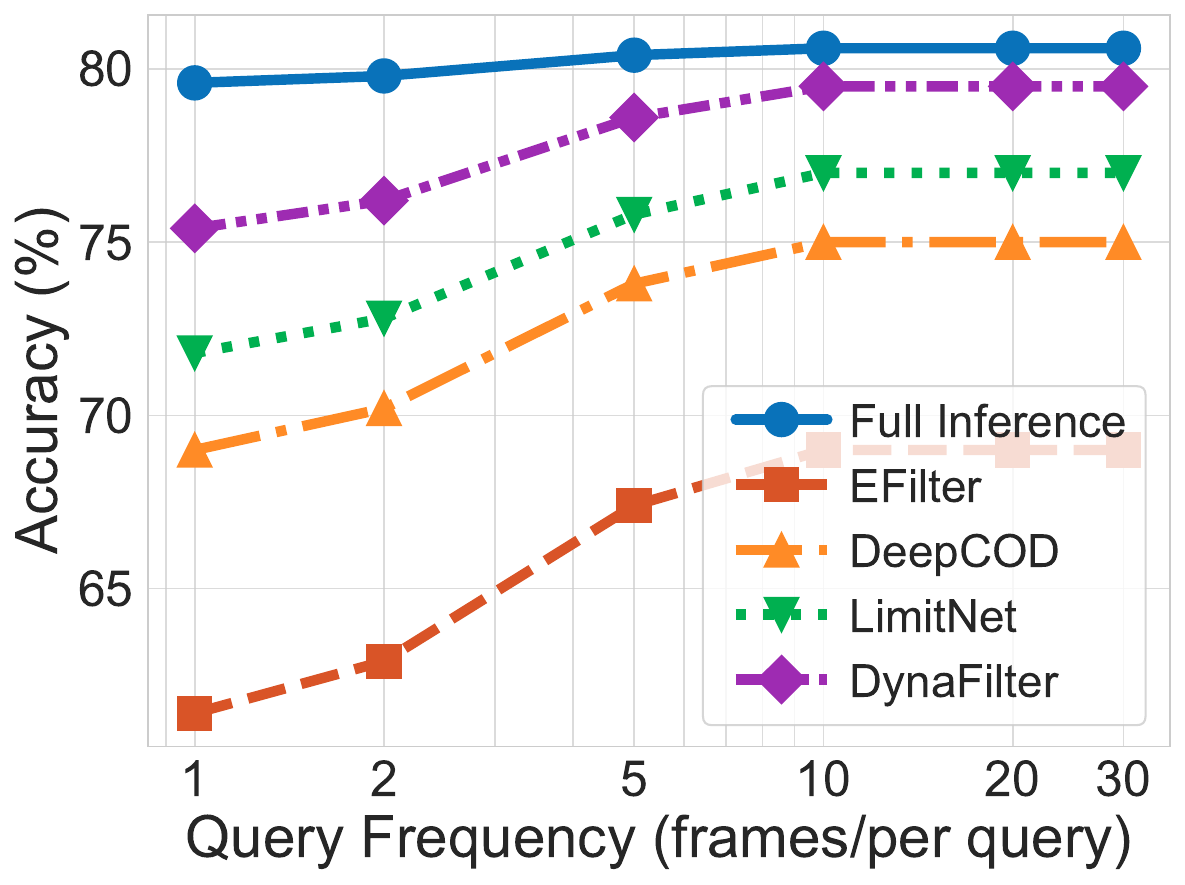}
\end{minipage}}
\subfigure[Inference latency (ms)]{
\begin{minipage}[b]{0.485\linewidth}
\includegraphics[width=1\linewidth]{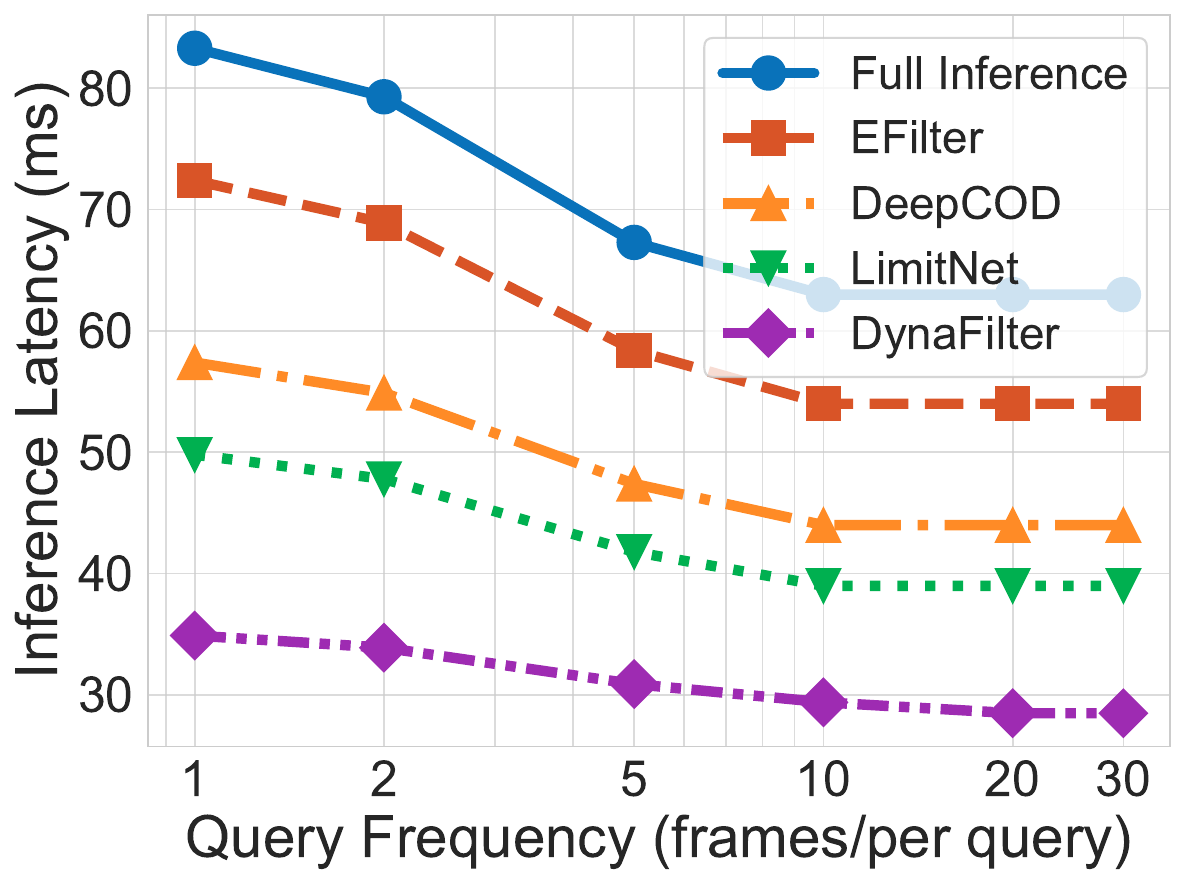}
\end{minipage}}
\subfigure[Object switching latency (ms)]{
\begin{minipage}[b]{0.485\linewidth}
\includegraphics[width=1\linewidth]{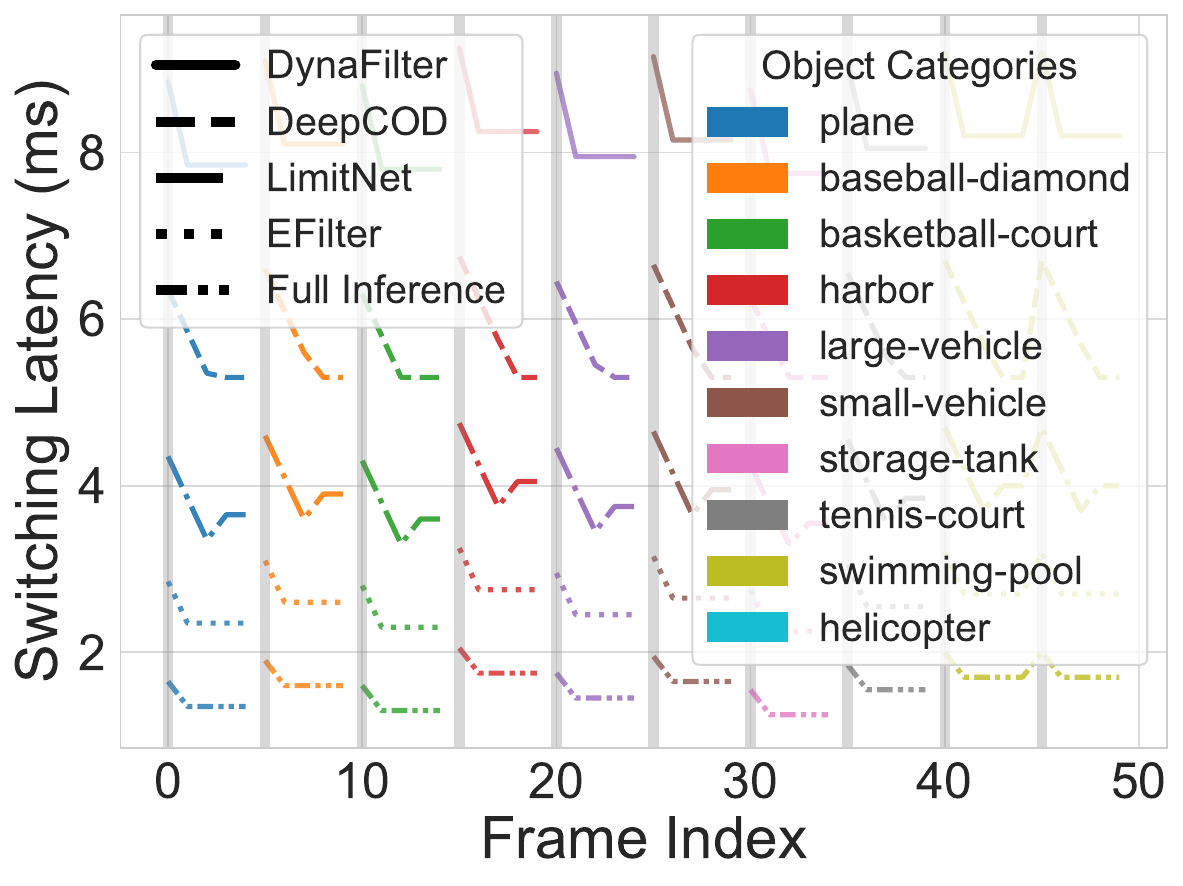}
\end{minipage}}
\caption{The impact of dynamic query filtering on the DOTA-v1.0 dataset.}
\Description{The impact of dynamic query filtering on the DOTA-v1.0 dataset.}
\vspace{-0.2cm}
\label{fig:dynamic}
\end{figure}

Figure~\ref{fig:dynamic} demonstrates the system's adaptability on DOTA-v1.0. DynaFilter achieves a 3.1$\times\sim$7.7$\times$ improvement in adaptation latency (Figure~\ref{fig:dynamic}(a)), due to its compact cloud-driven configurations. It maintains high accuracy even at high query frequencies (Figure~\ref{fig:dynamic}(b)) via robust multimodal feature fusion. Furthermore, inference latency remains significantly lower than the baselines (Figure~\ref{fig:dynamic}(c)), and object switching latency is minimized (Figure~\ref{fig:dynamic}(d)), confirming DynaFilter’s effectiveness in handling data drift.

\subsection{JPEG Quality and RoI Locator}
\begin{figure}
\setlength{\abovecaptionskip}{0pt}
\setlength{\belowcaptionskip}{0pt}
\centering
\subfigure[DOTA-v1.0 dataset]{
\begin{minipage}[b]{0.475\linewidth}
\includegraphics[width=1\linewidth]{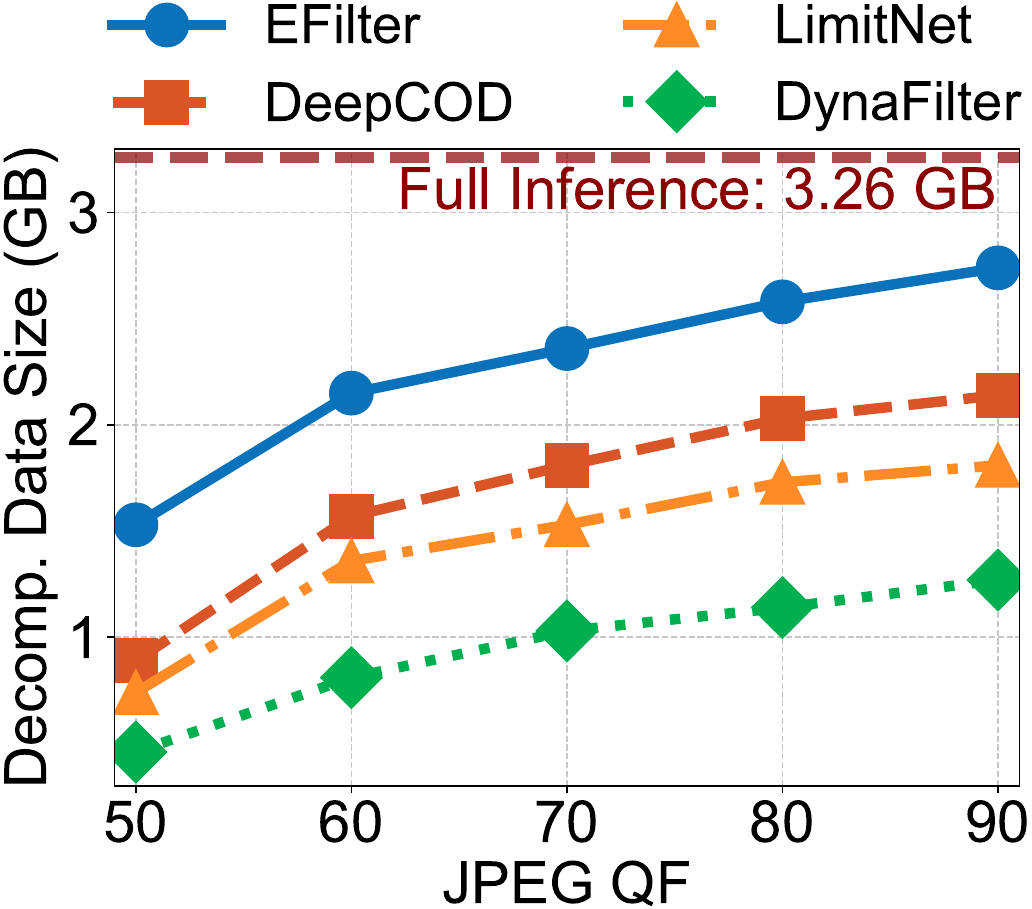}
\end{minipage}}
\subfigure[VisDrone 2019 dataset]{
\begin{minipage}[b]{0.495\linewidth}
\includegraphics[width=1\linewidth]{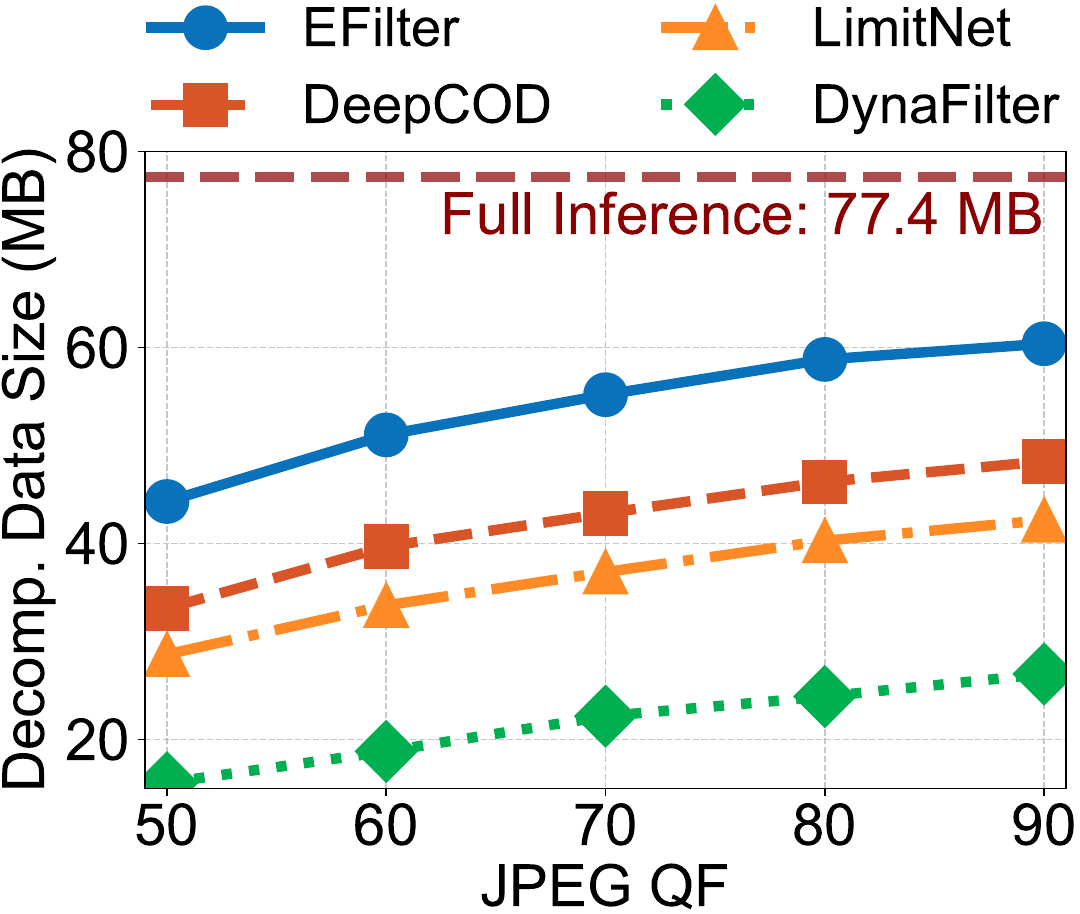}
\end{minipage}}
\caption{Decompressed data size with different JPEG quality factors (QFs) across two datasets.}
\Description{Decompressed data size with different JPEG quality factors (QFs) across two datasets.}
\vspace{-0.2cm}
\label{fig:decompression_comparison}
\end{figure}

We use decompressed data size for validating bandwidth efficiency across varying JPEG quality factors (QFs). As shown in Figure~\ref{fig:decompression_comparison}, DynaFilter reduces decompressed data size by 2.6$\times\sim$7.1$\times$ on DOTA-v1.0 and 2.9$\times\sim$5.0$\times$ on VisDrone compared to Full Inference. 
Lower QFs actually enhance filtering effectiveness as coding artifacts distinguish background from RoIs. DynaFilter outperforms DeepCOD ($1.8\times\sim2.5\times$) and LimitNet ($1.6\times\sim2.2\times$) by avoiding the transmission of irrelevant compressed features or full frames.

\begin{figure}
\vspace{-0.2cm}
\setlength{\abovecaptionskip}{0pt}
\setlength{\belowcaptionskip}{0pt}
\centering
\subfigure[RoI accuracy and latency]{
\begin{minipage}[b]{0.485\linewidth}
\includegraphics[width=1\linewidth]{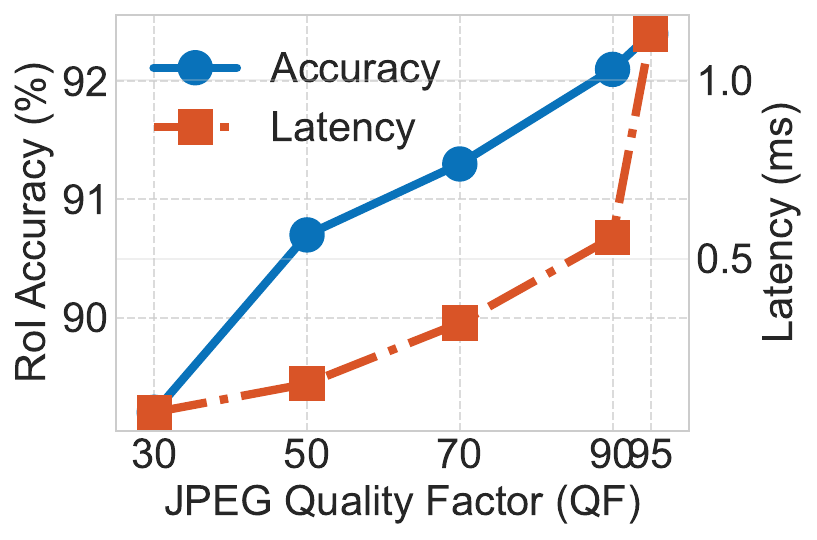}
\end{minipage}}
\subfigure[Memory and energy]{
\begin{minipage}[b]{0.485\linewidth}
\includegraphics[width=1\linewidth]{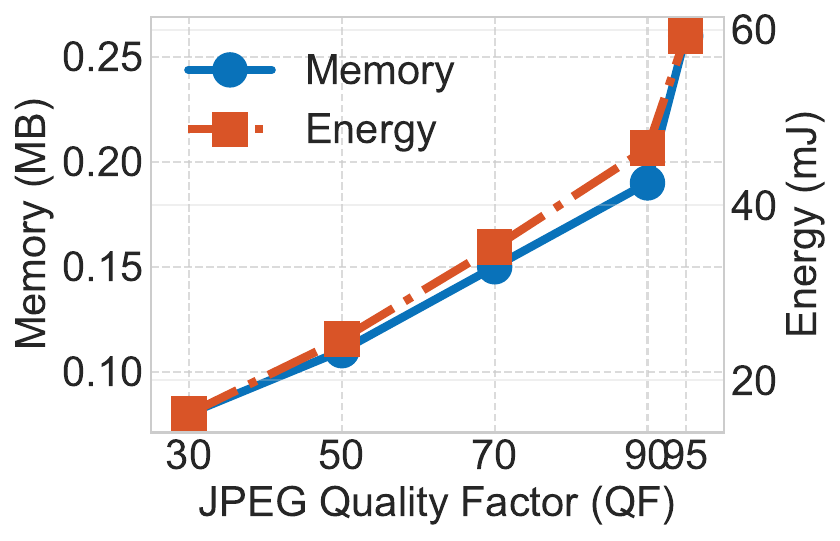}
\end{minipage}}
\caption{Performance of RoI Locator across different JPEG quality factors.}
\Description{Performance of RoI Locator across different JPEG quality factors.}
\vspace{-0.2cm}
\label{fig:JPEG_QF}
\end{figure}

Figure~\ref{fig:JPEG_QF} shows the performance of the RoI Locator across different JPEG QFs. The results show that the locator maintains consistent performance across the entire quality spectrum, with only minor variations in accuracy and memory overhead, while incurring negligible latency and energy consumption compared to Fill Inference. While lower QF settigs result in more aggressive quantization, the fundamental patterns that distinguish object regions from background are preserved, enabling the analyzer to effectively identify RoIs.

\subsection{Multimodal Feature Fusion Filter}
We evaluate the contributions of each feature modality to multi-object tracking accuracy (MOTA) for the object tracking task, with special attention to challenging scenarios including occlusion and small object tracking.

\begin{figure}
\setlength{\abovecaptionskip}{0pt}
\setlength{\belowcaptionskip}{0pt}
\centering
\subfigure[MOTA under small objects]{
\begin{minipage}[b]{0.45\linewidth}
\includegraphics[width=.95\linewidth]{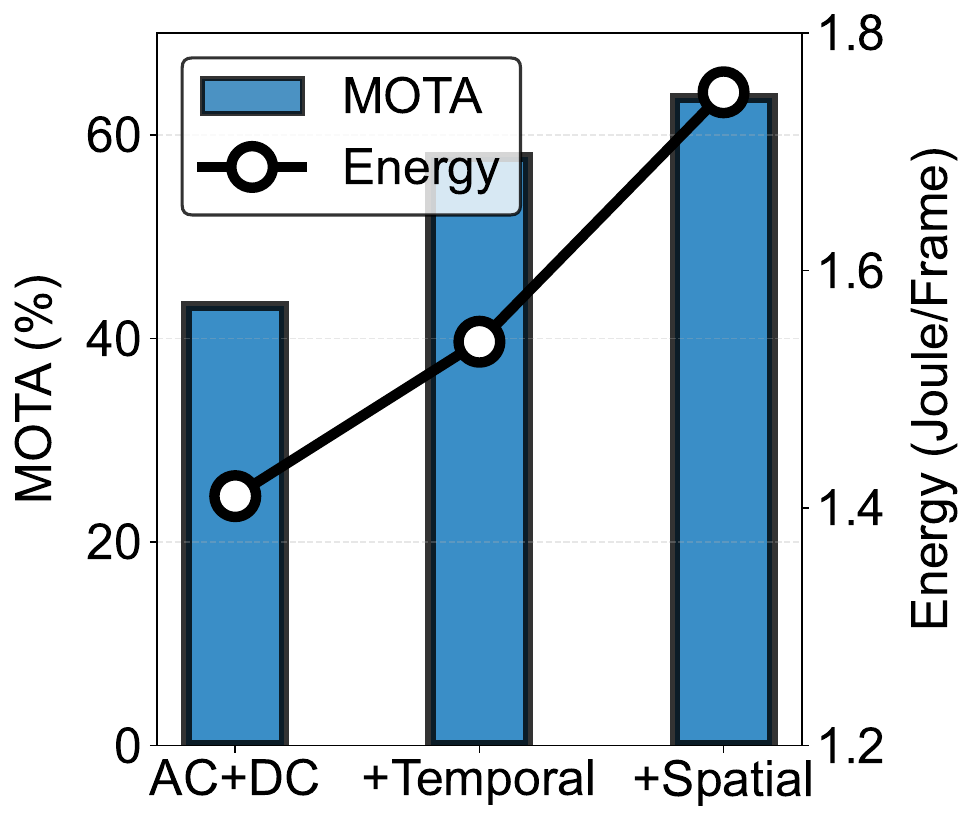}
\end{minipage}}
\subfigure[MOTA under occlusion]{
\begin{minipage}[b]{0.45\linewidth}
\includegraphics[width=.95\linewidth]{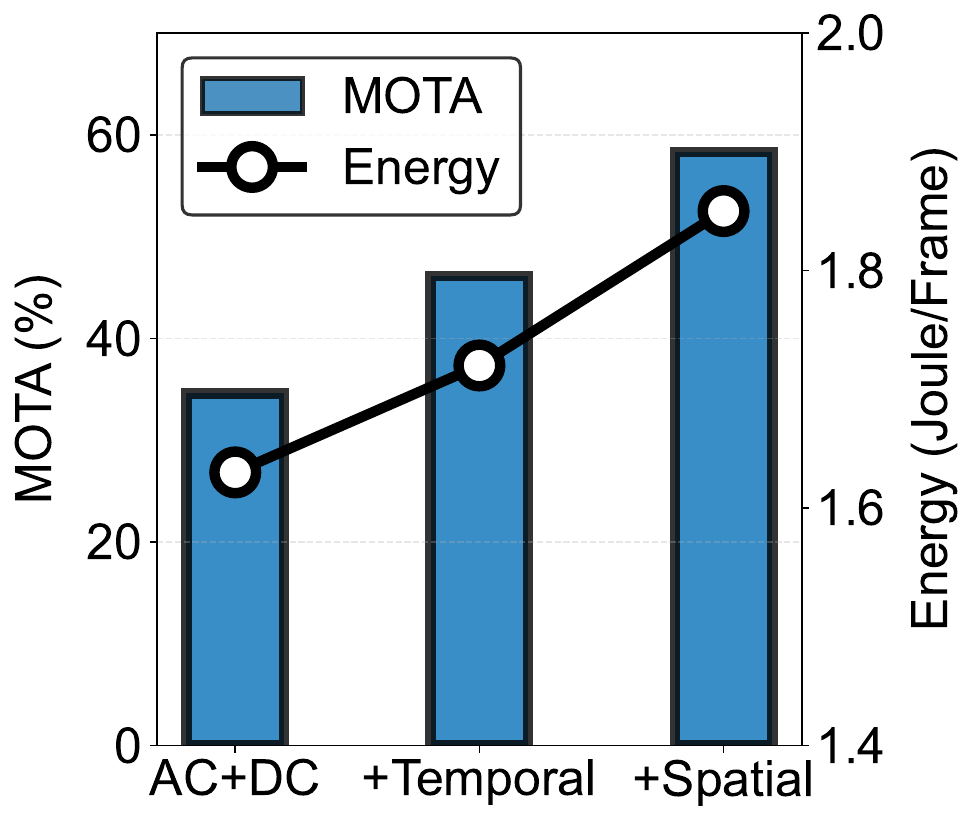}
\end{minipage}}
\caption{Performance comparison of multimodal feature fusion filter with different feature modality.}
\Description{Performance comparison of multimodal feature fusion filter with different feature modality.}
\vspace{-0.2cm}
\label{fig:MOTA}
\end{figure}

In Figure~\ref{fig:MOTA}, for objects smaller than 32×32 pixels, com\-pres\-sed-domain features alone achieve only 43.2\% MOTA due to weak and easily confused signatures in the compressed-domain. In comparison, temporal context features provide a dramatic 14.6\% improvement by capturing motion signatures that persist despite limited spatial footprint. Spatial location modeling contributes an additional 5.8\% MOTA improvement through Kalman filtering, which maintains trajectory consistency during brief visibility gaps.

For occlusion, compressed-domain features alone struggle during occlusion, with MOTA dropping to 34.6\% as objects temporarily disappear. Temporal context features provide a 11.5\% MOTA improvement by identifying objects through their motion history, while spatial location modeling contributes 12.2\% by predicting object positions during occlusion periods. For severe occlusion, the system increases emphasis on temporal continuity, while during motion blur, it shifts to spatial consistency.

\subsection{Enhancement Strategy}
We evaluate the impact of different strategies on the performance of the object tracking tasks, with comprehensive metrics including MOTA, tracking latency accounting for detection latency and correlation matching latency, and memory usage. 
The temporal-spatial feature enhancement strategy significantly improves tracking robustness by incorporating motion intensity analysis and spatial prediction. The tracking association enhancement strategy addresses the critical challenges of target occlusion and ID switching. 

\begin{figure}[tb]
\large
\centerline{\includegraphics[width=.9\linewidth]{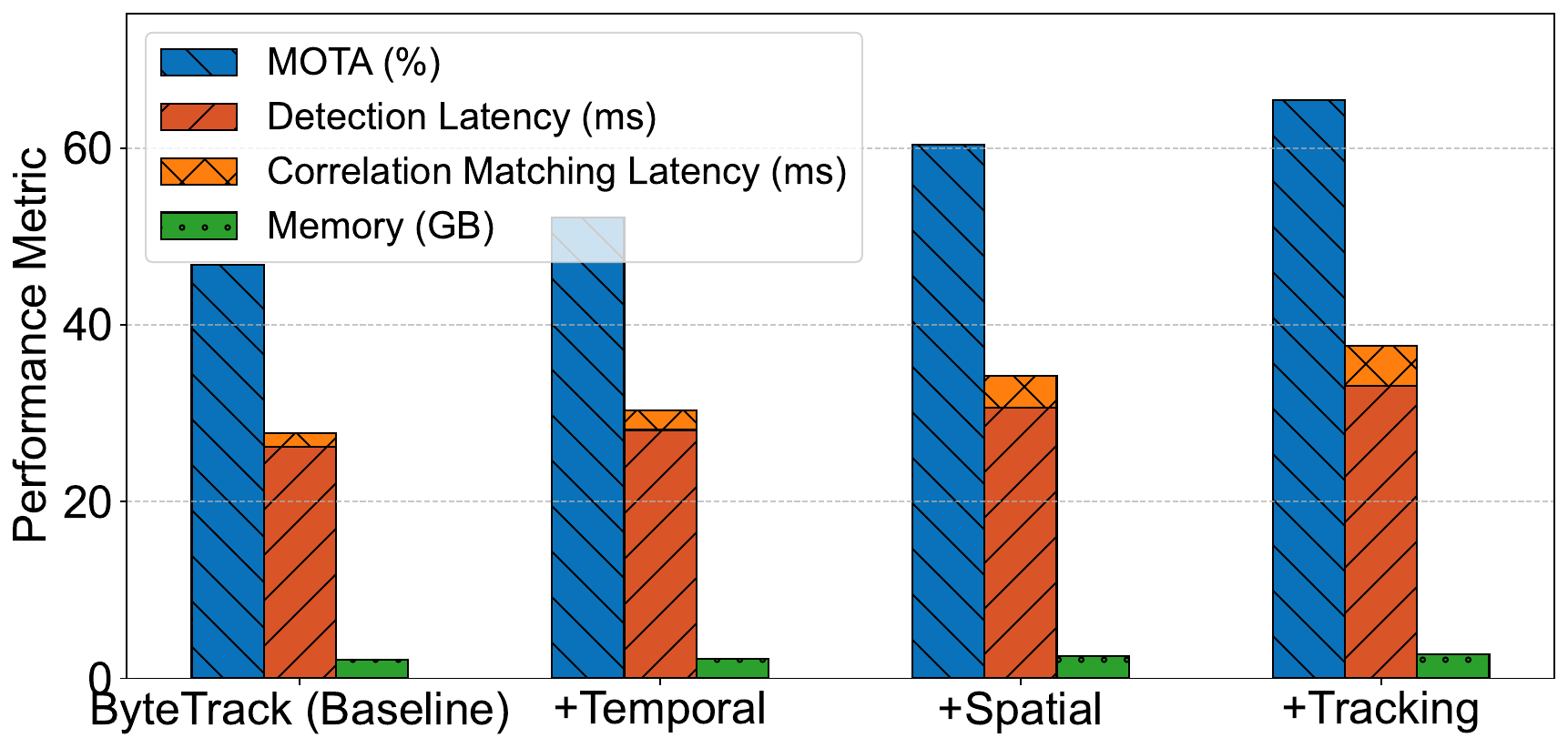}}
\vspace{-0.3cm}
\caption{Performance of RoI-based object tracking with different enhancement strategies.}
\Description{Performance comparison of RoI-based YOLOv8 for object tracking with different enhancement strategies.}
\vspace{-0.3cm}
\label{fig:ob}
\end{figure}

As shown in Figure~\ref{fig:ob}, the combination of the two enhancement strategies achieves significant performance improvements across all metrics. DynaFilter achieves a MOTA of 65.4\%, while the ByteTrack~\cite{zhang2022bytetrack} achieves a MOTA of 46.8\%. Meanwhile, the tracking latency and memory usage only increase by 26.3\% and 21.6\%, respectively.

\subsection{Video Streaming}
\begin{figure}
\setlength{\abovecaptionskip}{0pt}
\setlength{\belowcaptionskip}{0pt}
\centering
\subfigure[E2E latency and MOTA(\%)]{
\begin{minipage}[b]{0.485\linewidth}
\includegraphics[width=1\linewidth]{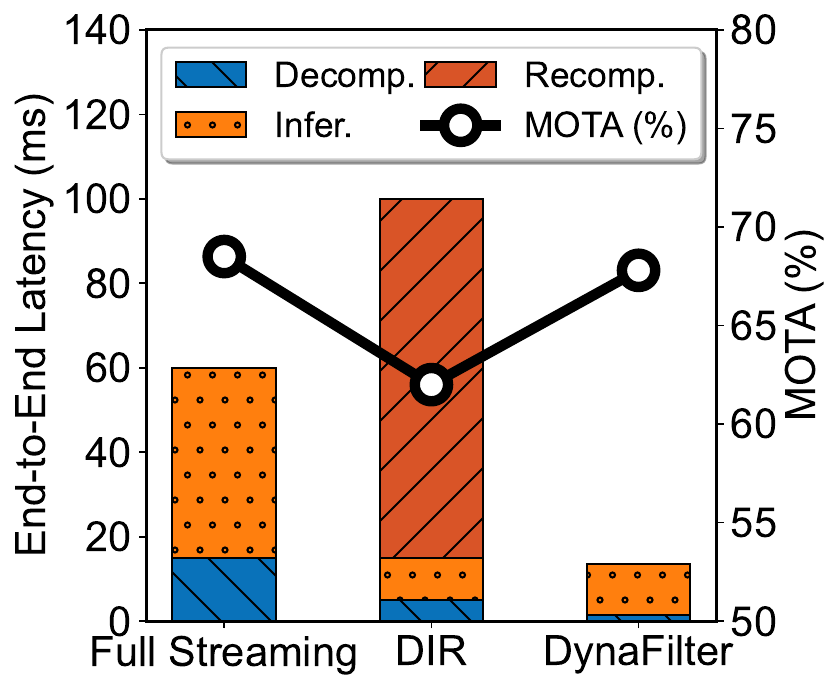}
\end{minipage}}
\subfigure[Energy and bandwidth saving]{
\begin{minipage}[b]{0.485\linewidth}
\includegraphics[width=1\linewidth]{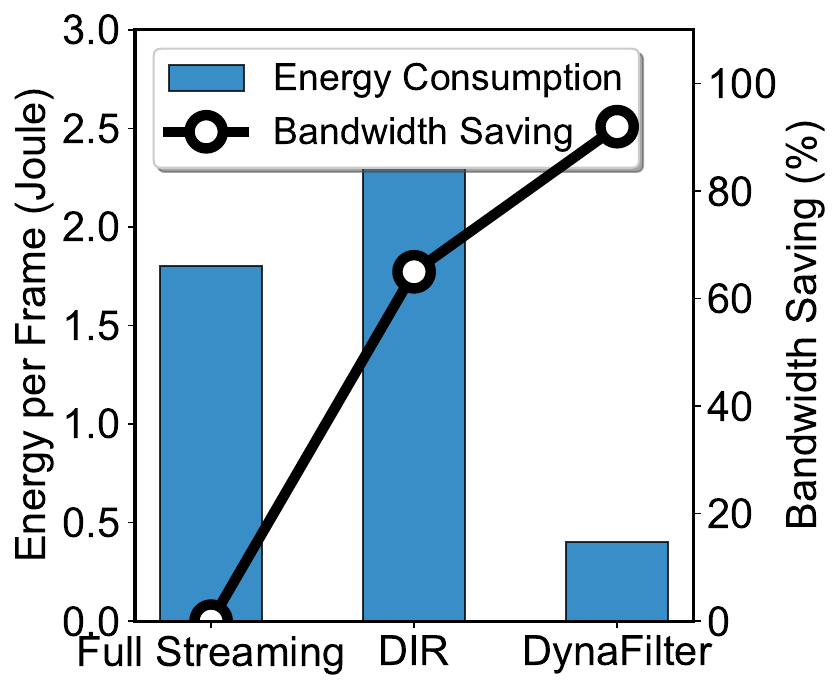}
\end{minipage}}
\caption{Performance comparison of JPEG/video codecs profiler for video streaming.}
\Description{Performance comparison of JPEG/video codecs profiler for video streaming.}
\vspace{-0.3cm}
\label{fig:video}
\end{figure}

Figure~\ref{fig:video} shows that DIR suffers high latency, about 100ms, due to the recompression bottleneck. In contrast, DynaFilter achieves 13.5ms latency, corresponding to a 7.4$\times$ speedup over DIR, by leveraging bitstream features. DynaFilter maintains 67.8\% MOTA, matching Full Streaming (68.5\%) and outperforming again DIR (62.0\%).

As for energy consumption, DynaFilter consumes just 0.4 J/frame, which is 83.3\% lower than DIR (2.4 J/frame), and achieves 92\% bandwidth savings by transmitting only query-relevant partial bitstreams.

\subsection{System Overhead}
\begin{figure}[tb]
\large
\centerline{\includegraphics[width=1\linewidth]{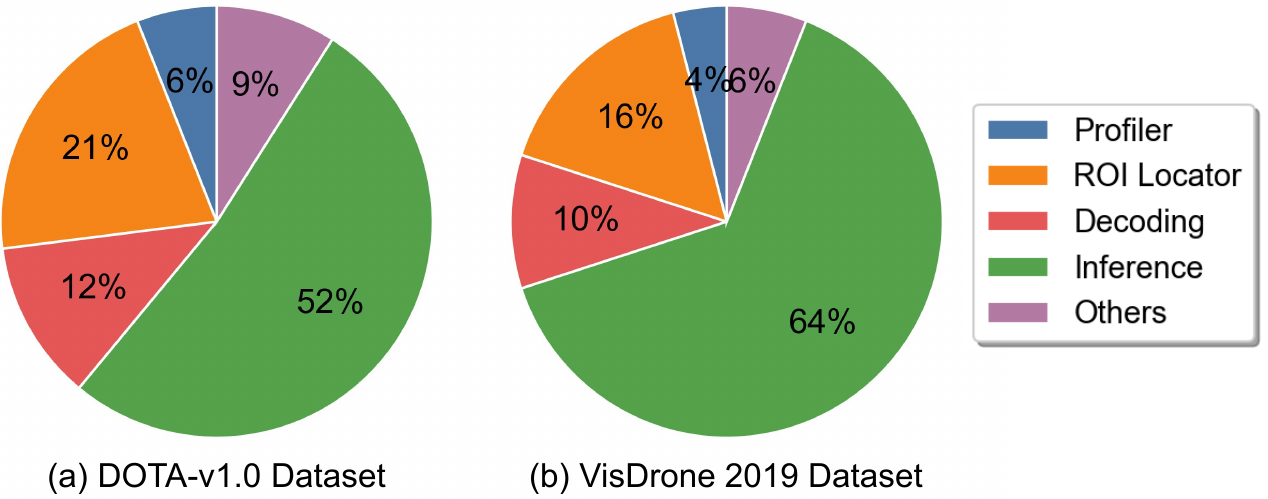}}
\caption{Execution latency of each component.}
\Description{Execution latency of each component.}
\vspace{-0.2cm}
\label{fig:time}
\end{figure}

We use 1080p video input for  breaking down the different latency factors on Jetson Orin Nano.
As shown in Figure~\ref{fig:time}, DynaFilter's processing pipeline demonstrates exceptional time efficiency. Specifically, inference remains the most computationally intensive component (52\% for DOTA-v1.0, 64\% for VisDrone), while our innovative compressed-domain filtering components contribute minimally to overall latency. The profiler accounts for only 4$\sim$6\% of processing time, and the RoI Locator adds just 16$\sim$21\%, demonstrating the lightweight nature of our compressed-domain analysis.

\begin{figure}[tb]
\vspace{-0.2cm}
\large
\centerline{\includegraphics[width=1\linewidth]{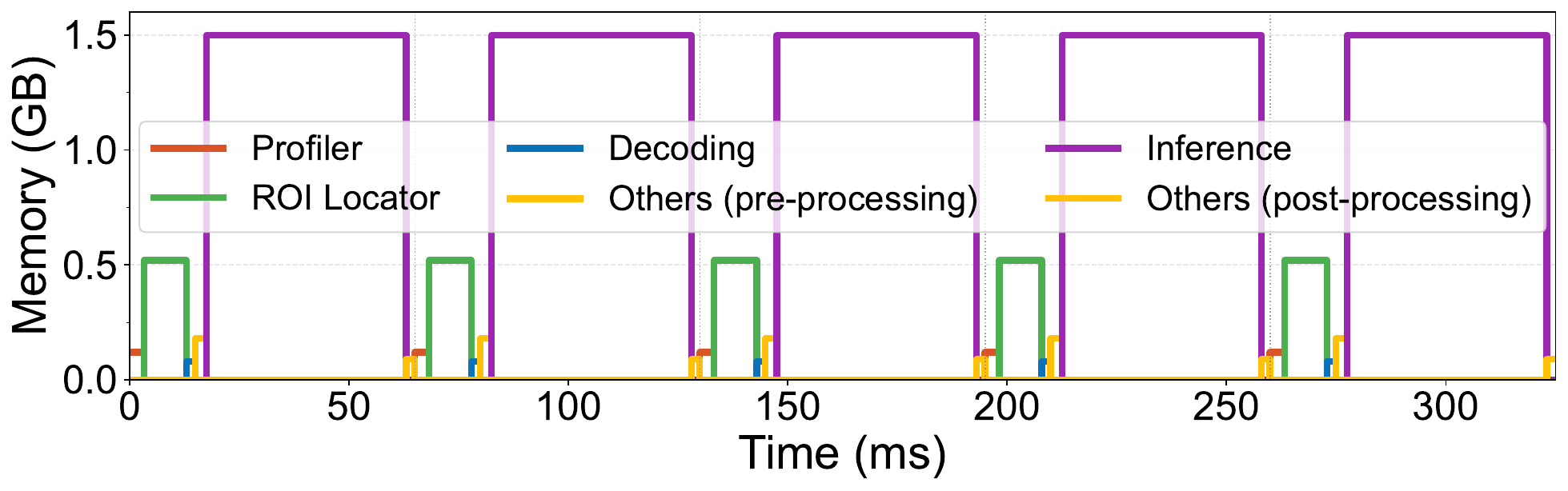}}
\vspace{-0.3cm}
\caption{Memory usage of the DynaFilter pipeline.}
\Description{Memory usage of the DynaFilter pipeline.}
\vspace{-0.5cm}
\label{fig:memory}
\end{figure}

As for memory consumption, Figure~\ref{fig:memory} reveals that DynaFilter achieves a peak usage of only 2.49GB, representing a 36\% reduction compared to full inference's 3.87GB. By processing only the identified RoIs, DynaFilter reduces the memory required for the inference stage. 
The consistent memory profile of DynaFilter (±0.12GB across different scene types) ensures system robustness. 

\section{Related work} \label{Related}

Our work intersects different areas, whose existing literature we compare with our efforts next.

\fakepar{Computational offloading}
Recent advances focus on computational offloading strategies~\cite{yao2020deep,huang2022real,hu2020starfish,zhang2024dvfo} to address resource constraints of edge devices. 
DeepCOD~\cite{yao2020deep} and LimitNet~\cite{hojjat2024limitnet} design encoders with different architectures to achieve efficient offloading.
Elf~\cite{zhang2021elf} accelerates high-resolution visual processing by partitioning video frames and offloading them to multiple servers in parallel. 
AgileNN~\cite{huang2022real} and E3~\cite{lin2025e3} uses explainable AI to achieve on-device inference.
However, these works cannot address data drift.
DynaFilter introduces a cloud driven-edge inference paradigm, enabling edge devices to perform filtering directly in the compressed-domain without full decompression.

\fakepar{Compressed-domain processing}
Recent advances focus on compressed-domain processing using deep learning. 
For example, CoVA~\cite{hwang2022cova} splits video frames between compressed and pixel domains to address decoding bottlenecks.
MFCD-Net~\cite{battash2020mimic} uses imitation learning to mimic raw domain networks in the compressed-domain for action recognition.
CD-VSR~\cite{chen2021compressed} enhances video quality directly in the compressed-domain.
EBM~\cite{xing2024enhancing} optimizes enhancement bias towards com\-pres\-sed-domain to improve the quality of compressed images.
However, these works treat compression as a bottleneck, rather than a computational medium.
DynaFilter fundamentally differs by establishing a precise mapping between semantic parameters and compressed-domain features.

\fakepar{Dynamic filtering}
Recent advances explore various approaches for dynamic filtering. 
EfficientVIS~\cite{wu2022efficient} uses temporal information to improve segmentation accuracy, but cannot adapt to changing queries in the cloud. 
Meta-Filter~\cite{xu2021learning} predicts position- and channel-specific filter weights for few-shot learning. 
SQ-MG~\cite{zhang2021accurate} presented an object detection framework with query-support mutual guidance, but operates exclusively in the pixel domain. Reducto~\cite{li2020reducto} implements frame-level filtering using low-level video features, but focuses on temporal filtering.
However, these works treat filters as static, meaning they need to be retrained or reanalyzed when queries change. In contrast, DynaFilter enables the cloud to dynamically generate new filter configurations based on changing queries.

\section{Discussion} \label{Discussion}

We articulate limitations of our work and possible extensions.

\fakepar{Extension to modern and neural codecs}
The landscape of image and video compression is evolving toward modern formats (e.g., WebP, AVIF, HEIC) and emerging neural video compression. Our core insight, leveraging the correlation between compression syntax and high-level semantics, remains applicable to these standards. For block-based formats like AVIF, DynaFilter can be adapted to profile its specific partition structures and frequency transforms. For neural codecs that compress data into latent feature representations, we may extend DynaFilter to perform filtering directly in the latent space.

\fakepar{Adaptive filter generation}
The filter configurations of DynaFilter rely on historical data. We can integrate online learning~\cite{hoi2021online} to enable adaptation based on immediate feedback from edge devices. One promising approach involves a federated meta-learning~\cite{fallah2020personalized}.
For instance, integrating model-agnostic meta-learning~\cite{finn2017model} could enable the cloud to rapidly adapt filter configurations to new object categories. 
Additionally, developing neural architecture search (NAS)~\cite{cai2018proxylessnas} specifically for filter generation could optimize the trade-off between accuracy and computational overhead. 

\fakepar{Cross-modal extension}
The cloud driven-edge execution paradigm of DynaFilter is fundamentally applicable to any bandwidth-constrained multimodal application, such as underwater exploration or remote sensing. Our principle of mapping high-level semantics to low-level compression features could process diverse data types with consistent efficiency by abstracting them into a unified semantic feature space.
We can develop cross-modal learning~\cite{ouyang2022cosmo} that enables knowledge transfer between different data types. 
For instance, contrastive learning~\cite{chuang2020debiased,khosla2020supervised} could be employed to learn unified representations across modalities, allowing the system to leverage complementary information from different sensor types to improve overall accuracy.

\section{Conclusion}\label{Conclusion}
This paper proposes DynaFilter, a cloud-driven technique for dynamic filtering. By accurately mapping low-level com\-pres\-sed-domain features with high-level semantics, DynaFilter can identify RoI without full decompression. 
Extensive experiments show that DynaFilter delivers significant
improvement over state-of-the-art baselines in decompressed data size, energy consumption and inference latency.

\newpage

% \section*{Acknowledgment}
% We thank our anonymous reviewers and shepherd for the helpful comment and feedback. This work is partly supported by.

% \bibliographystyle{unsrt}
\balance
\bibliographystyle{ACM-Reference-Format}
\bibliography{ref}
% \end{sloppypar}

\end{document}